\pdfoutput=1
\documentclass[11pt]{article}
\usepackage{bbm}
\usepackage{comment}
\usepackage{multirow}

\usepackage{physics}
\usepackage{acl}
\usepackage{array}
\usepackage{color,soul}
\usepackage{placeins}
\usepackage{amssymb}
\usepackage{booktabs, multirow, array, makecell, caption}

\usepackage{xcolor}
\usepackage{times}

\usepackage{latexsym}
\usepackage[ruled,linesnumbered]{algorithm2e}
\RestyleAlgo{ruled}
\usepackage{tabularx}
\usepackage{hyperref}
\usepackage{pifont}
\usepackage{verbatim}
\usepackage{soul}

\usepackage{subcaption}
\usepackage{mwe}

\usepackage{mathtools}
\usepackage{lipsum}
\usepackage{float}
\newcolumntype{b}{X}
\newcolumntype{s}{>{\hsize=.5\hsize}X}

\usepackage[T1]{fontenc}

\usepackage[utf8]{inputenc}

\usepackage{microtype}

\title{PCC: Paraphrasing with Bottom-k Sampling and Cyclic Learning
for \\ Curriculum Data Augmentation\thanks{\,\,\,The work described in this paper is substantially supported by a grant from the Research Grant Council of the Hong Kong Special Administrative Region, China (Project Code: 14200719).}}

\author{Hongyuan Lu, Wai Lam\\
  The Chinese University of Hong Kong \\
  \texttt{\{hylu,wlam\}@se.cuhk.edu.hk} \\}

\begin{document}
\maketitle
\begin{abstract}
Curriculum Data Augmentation (CDA) improves neural models by presenting synthetic data with increasing difficulties from easy to hard. However, traditional CDA simply treats the ratio of word perturbation as the difficulty measure and goes through the curriculums only once. This paper presents \textbf{PCC}: \textbf{P}araphrasing with Bottom-k Sampling and \textbf{C}yclic Learning for \textbf{C}urriculum Data Augmentation, a novel CDA framework via paraphrasing, which exploits the textual paraphrase similarity as the curriculum difficulty measure. We propose a curriculum-aware paraphrase generation module composed of three units: a paraphrase candidate generator with bottom-k sampling, a filtering mechanism and a difficulty measure. We also propose a cyclic learning strategy that passes through the curriculums multiple times. The bottom-k sampling is proposed to generate super-hard instances for the later curriculums. Experimental results on few-shot text classification as well as dialogue generation indicate that PCC surpasses competitive baselines. Human evaluation and extensive case studies indicate that bottom-k sampling effectively generates super-hard instances, and PCC significantly improves the baseline dialogue agent.
\end{abstract}
\section{Introduction}
Data augmentation techniques create artificial data mixed with the original data for improved performance. Traditional data augmentation techniques in the language community include word-level perturbation such as synonym replacement, random insertion, random swap, and random deletion \citep{wei-zou-2019-eda}. Sentence-level techniques such as Round-trip Translation \citep{sennrich-etal-2016-improving} exploits the use of machine translation models to translate the input sentence to another language before translating back to the source language which can be essentially treated as a form of paraphrasing.
\par
Curriculum learning presents training instances in a meaningful order with increasing difficulties to neural models for a boost in performance. Traditional curriculum learning \citep{BengioLCW09, ijcai2018-587,ijcai2020-534,platanios-etal-2019-competence, xu-etal-2020-curriculum,xu-etal-2020-dynamic, su-etal-2021-dialogue} categorizes the original training instances into different levels of difficulties to be gradually presented to the model where a core component called difficulty measure, which is usually defined as a numerical number where a bigger number indicates a more difficult sample. 
\par
Combining the merits of the above two mentioned techniques, Curriculum Data Augmentation (CDA) creates synthetic data with increasing levels of difficulties to be presented to our neural models. Existing CDA defines the ratio of the words perturbation as the difficulty measure for curriculums and a gradual course which increases the difficulty of curriculums when the training loss plateaus \citep{wei-etal-2021-shot}, which then ends when the most challenging curriculum ends. Although
existing CDA is effective, yet there are several disadvantages. First, it employs word-level perturbation. This superficial operation keeps the augmentation to have a similar sentence structure as the original one. Next, it employs random insertion, random swap, and random deletion for augmentation. Although this can be durable as for text classification \citep{wei-etal-2021-shot}, this is not suitable for generation tasks, particularly when many words are perturbed, which can even easily break the sentence grammar. Third, it uses a gradual course that only enters each level of difficulty once. A typical problem in neural network training called catastrophic forgetting \citep{Kirkpatrick3521} can potentially happen in such a course, where the model might undesirably gradually forget some early learned knowledge.
\par
To mitigate the problems of word-level perturbation, we propose that paraphrasing \citep{yang-etal-2021-contrastive-representation} can be a source of data augmentation, which provides diverse and grammatically correct augmentation. However, it is non-trivial to utilize paraphrase augmentation in a curriculum setting.  Inspired by the fundamental linguistic concept of mutual implication \citep{10.2307/4320465, 10.2307/20013377}, we treat two sentences as a pair of paraphrases if they can infer each other. For example, `I am glad to help you.' and `Let me help you out!' can be a pair of paraphrases, which provides a diverse change of the sentence structure suitable for the curriculum setting. We also employ textual similarity for our difficulty measures for the curriculum. Higher scores indicate that two sentences are more textually similar to each other. Specifically, we treat pairs with lower scores as more difficult instances to be presented in later curriculums. We propose a paraphrase candidate generator integrated with bottom-k sampling.
Traditional sampling methods such as top-k sampling \citep{topk} and top-p \citep{topp} sampling tend to generate easier paraphrases that have relatively high similarity scores. We propose bottom-k sampling to generate super-hard paraphrases for the later harder curriculums by pruning the most probable words.\footnote{Note that we still use a combination of top-k and top-p sampling for generating easier curriculums.} This leads the generation towards a more grammatically and lexically diverse paraphrase sampling space with low textual similarity.
\par
To mitigate catastrophic forgetting, we propose to incorporate cyclic learning to pass through the curriculums multiple times.
\par
In summary, our proposed framework, called \textbf{PCC}: \textbf{P}araphrasing with Bottom-k Sampling and \textbf{C}yclic Learning for \textbf{C}urriculum Data Augmentation, makes three contributions:
\begin{itemize}
\setlength\itemsep{0em}
    \item We exploit the use of paraphrasing with mutual implication as a data augmentation source in curriculum learning.
    \item To generate mutual implicative paraphrases, we propose a curriculum-aware paraphrase generation module composed of three units, namely, a paraphrase candidate generator with bottom-k sampling for generating super-hard instances, a filtering mechanism, and a difficulty measure using textual similarity.
    \item We propose cyclic learning to enter each curriculum multiple times.
\end{itemize}
Experimental results indicate that PCC surpasses competitive baselines on few-shot text classification as well as dialogue generation. Human evaluation indicates that bottom-k sampling effectively generates grammatically and lexically rich paraphrases, and PCC significantly improves our baseline dialogue agent. To our best knowledge, this is the first time to apply CDA on a generation task.
\par 
\paragraph{Takeaway} Overall, we present the effectiveness of paraphrasing as a curriculum data augmentation technique. The use of cyclic learning and bottom-k sampling further boosts performance. With some modifications, future works can treat PCC as a data augmentation framework and adapt it to other downstream tasks. Future works can also leverage bottom-k sampling in generating textual outputs that are grammatically and lexically rich.
\section{Related Work}
\subsection{Data Augmentation}
Existing textual data augmentation techniques can be broadly categorized into two streams: word-level and sentence-level augmentation.
\par
For word-level augmentation, well-known operations includes synonym replacement \citep{10.5555/2969239.2969312}, random insertion, random deletion and random swap \citep{wei-zou-2019-eda}. In contrast to dictionary-based synonym replacement, another stream of works randomly replace words with masks and employs BERT models for predicting the words as a source of augmentation that exploits the contexts \citep{Wu2019ConditionalBC, inproceedings123}.
\par
For sentence-level augmentation, Round-trip Translation \citep{sennrich-etal-2016-improving} augments translation pairs by translating from the source language into the target language, and back to the source language with two machine translation models. \citet{gao-etal-2020-paraphrase} proposes to use paraphrases as a source of augmentation in task-oriented dialogue generation. It has also been proposed to retrieve from unpaired corpora as a source of augmentation in the dialogue community \citep{zhang-etal-2020-dialogue}. Another stream of work edits the retrieved dialogue response for better generation \citep{cai-etal-2019-skeleton, cai-etal-2019-retrieval}, which can be treated as a form of indirect augmentation. The closest work to ours is \citet{gao-etal-2020-paraphrase}, where theirs does not employ curriculum learning.
\subsection{Curriculum Learning}
While traditional curriculum learning sorts the training samples in an order of increasing difficulties \citep{BengioLCW09, pmlr-v80-weinshall18a, su-etal-2021-dialogue}, our method follows the other stream of works that applies transformation on the original data with dedicated difficulty level \citep{10.5555/3327757.3327874,Ganesh, wei-etal-2021-shot}. The closest work to ours is \citet{wei-etal-2021-shot}. Their work does not consider paraphrasing and focuses on text classification only.
\\
\\
\SetKwInput{KwInput}{Input}
\SetKwInput{KwOutput}{Output}
\SetAlCapNameFnt{\small}
\SetAlCapFnt{\small}
\SetKwComment{Comment}{/* }{ */}
\begin{algorithm}[t!]
\small
\small {\caption{Paraphrasing with Bottom-k Sampling and Cyclic Learning for Curriculum Data Augmentation (PCC)} \label{alg:pcc}}
\KwInput{Dataset $\mathcal{D}$ for the downstream task;}
\KwOutput{Trained downstream task model;}
For the entire dataset $\mathcal{D}$, invoke the curriculum-aware paraphrase generation module with $\mathcal{D}$ and cache the augmentation results $\bar{\mathcal{D}}$ for training purpose\;
\While{not the end of training}{
   Set difficulty level $l$ to $0$ at the start of a cycle\;
   \While{not the end of current cycle}{
   \While{not the end of current curriculum}{
       Uniformly sample the next batch of training instance $\mathcal{S}$\; 
       Invoke the curriculum-aware paraphrase generation module for each training instance in $\mathcal{S}$ to retreive a batch of training augmentation $\mathcal{T}$ with difficulty level $l$.\;
       Invoke the task-specific model trainer to train the downstream task model with the training augmentation $\mathcal{T}$\;
   }
   Increase $l$ by $1$ to the next level at the end of current curriculum\;
   }
}
\end{algorithm}
\begin{algorithm}[t!]
\small
\small {\caption{Curriculum-aware Paraphrase Generation Module}\label{alg:pcc2}}
\KwInput{A single training instance with textual input $x$; difficulty level $l$;}
\KwOutput{Cache the generated paraphrases into $\bar{\mathcal{D}}$ or retrieve an augmented training instance $\bar{x}$;}
\uIf{a cached augmentation exists}
{Retrieve $\bar{x}$ that corresponds to $x$ with the difficulty measure $d=l$\;}
  \Else{
    Invoke the paraphrase candidate generator integrated with bottom-k sampling to generate a bag of paraphrase candidates for $x$\;
    Invoke the mutual implication classifier for each paraphrase candidate to obtain corresponding binary indicator against the input sentence\;
    Calculate the textual similarity for each paraphrase candidate against the input\;
    Filter the generated paraphrase candidates with the mutual implication and the textual similarity using Equation \ref{filtering}\;
    Assign a difficulty measure $d$ to the filtered paraphrases with Equation \ref{diff}\;
    Cache the augmentation results into $\bar{\mathcal{D}}$ \;
  }
\end{algorithm}
\vspace{-7mm}
\section{Our Proposed Framework}
\subsection{Background of Curriculum Data Augmentation (CDA)}
 Existing CDA \citep{wei-etal-2021-shot} varies the word-level perturbation ratio to achieve different levels of difficulties under curriculum learning with simple word perturbation strategies such as synonym replacement, random insertion, swap, and deletion. As illustrated in Figure \ref{curriculum_vs}, such simple word perturbation strategies create problematic instances that break the sentence grammar, which can hamper the model performance. There are two common CDA strategies. One is called two-stage curriculum, which uses a fixed perturbation ratio for a single curriculum as the second stage after training with the original data. The other one is called gradual curriculum. It uses different ratios for a number of (typically 5) curriculums with increasing difficulties. However, such a learning strategy ends after passing through all the curriculums only once, and catastrophic forgetting can happen.
 \subsection{Our Proposed PCC}
We propose curriculum data augmentation with paraphrase augmentation known as \textbf{P}araphrasing with Bottom-k Sampling and \textbf{C}yclic Learning for \textbf{C}urriculum Data Augmentation (PCC). Algorithm \ref{alg:pcc} depicts an overview of the whole PCC framework. At the start of training, we generate cached training augmentation for the entire dataset with our proposed curriculum-aware paraphrase generation module. Thereafter, we begin with the easiest curriculum. For each training instance, we retrieve the cached augmentation that has an equivalent difficulty measure with the current difficulty level. We then invoke the task-specific model trainer to train the downstream task model with the retrieved training augmentation. At the end of each curriculum difficulty level, we increase the difficulty level to advance to the next harder curriculum. In case it hits the end of the most difficult curriculum, we set the difficulty level to the easiest to start a new cycle. We propose such a cyclic learning strategy for mitigating potential catastrophic forgetting. In order to retrieve paraphrasing augmentation with appropriate difficulty measures, we propose a curriculum-aware paraphrase generation module.
\subsubsection{Curriculum-aware Paraphrase Generation Module}
Algorithm \ref{alg:pcc2} depicts the curriculum-aware paraphrase generation module. Three components are designed, namely, a paraphrase candidate generator integrated with a bottom-k sampling strategy, a filtering mechanism, and a difficulty measure. The paraphrase candidates are generated and then passed to the filtering mechanism. Finally, the filtered paraphrases are assigned a difficulty measure which represents to which curriculum difficulty level the augmentation belongs. 
\begin{figure}[t!]
\begin{center}
\vspace{0mm}
\centerline{
\includegraphics[width=7.5cm]{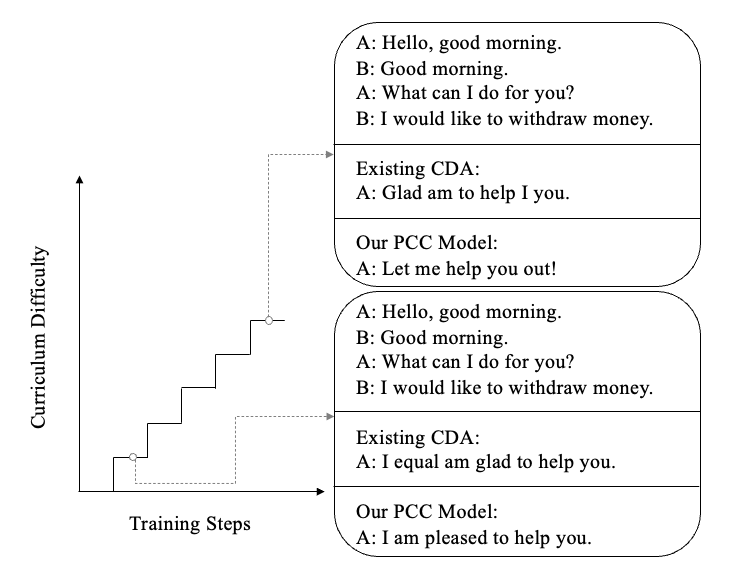}}
\caption{An illustrated example for our PCC model compared to existing CDA for dialogue generation. The original sentence is `I am glad to help you.'}
\label{curriculum_vs}
\end{center}
\vspace{-5mm}
\end{figure}
\begin{table}[t!]
\scriptsize
\centering
\begin{tabular}{ccc}
\begin{minipage}{\linewidth}
\begin{tabularx}{\textwidth}{s|X|s}
\hline
\noalign{\vskip 1mm} 
\textbf{Sample No.} & \textbf{Sample Text} & \textbf{Sim. Score}\\
\noalign{\vskip 1mm}    
\hline
\hline
\noalign{\vskip 1mm}  
1) & I am glad to assist you. & $0.888$\\
\noalign{\vskip 1mm}  
\hline
\noalign{\vskip 1mm}   
2) & Let's help you. I am glad to help you. & $0.619$ \\
\noalign{\vskip 1mm}  
\hline
\noalign{\vskip 1mm}   
3) & Thank you for contacting me. I am glad to help you. & $0.371$
\\
\noalign{\vskip 1mm}  
\hline \noalign{\vskip 1mm}  
4) & It is now my pleasure to help you. & $-0.038$ \\
\noalign{\vskip 1mm} 
\hline
\noalign{\vskip 1mm}  
5) & Let me help you out! & $-0.265$\\
\noalign{\vskip 1mm}  
\hline
\noalign{\vskip 1mm} 
6) & Thank you for your question. & $-0.506$\\
\noalign{\vskip 1mm}  
\hline \noalign{\vskip 1mm} 
\end{tabularx}
\end{minipage}
\end{tabular}
\caption{\label{MI_examples}
Paraphrases with mutual implication for an input `I am glad to help you.'
}
{\vskip -2mm}
\end{table}
\paragraph{Paraphrase Candidate Generator with Bottom-k Sampling} 
In order to generate mutual implicative paraphrases for the purpose of curriculum data augmentation, we adopt a Seq2Seq \citep{S2S} generator which receives an input sentence $x$ and generates the paraphrases $\bar{x}$ in an autoregressive manner \citep{nighojkar-licato-2021-improving}. During training, the paraphrase candidate generator is trained by maximising the following likelihood: \begin{align*}
    P\,(\bar{x}\mid x)=\prod_{t=1}^{T}P\,(\bar{x}_t\mid \bar{x}_1,..., \bar{x}_{t-1}, x),
\end{align*}
where $T$ represents the token length of the paraphrase and $x_{t}$ represents the word at the position $t$ that has been inferenced.
\par
Traditional sampling methods such as top-k sampling \citep{topk} and top-p sampling \citep{topp} sample the next token to be presented in the output from the most probable vocabularies that dominate the probability distribution. For example, at the \textit{i}-th timestep during inference, top-k sampling samples the next token $\bar{x}_i$ from the most probable $k$ words with the distribution:
\begin{equation}
\label{btk}
P_{\bar{x}_i\in \mathcal{V}^{(k)}}(\bar{x}_i\mid \bar{x}_1,..., \bar{x}_{i-1}, x),
\end{equation}
where $\mathcal{V}^{(k)}$ represents the most probable $k$ words. However, they are not suitable for generating super-hard instances, i.e., their output paraphrases tend to be textually similar to the original input sentence.\footnote{We found that top-k and top-p sampling tend to copy  dominating words from the input into the paraphrases. This is also the reason why we prefer bottom-k over bottom-p, as we would like to effectively prevent from coping dominating words. Appendix \ref{bottom-kcases} presents a detailed analysis.}
\par
To avoid coping the words and  unearth the super-hard paraphrases to be used in later curriculums, we propose bottom-k sampling\footnote{We give it such a name to make it catchy. It does not sample from the bottom $k$ words. It samples from the bottom $|\mathcal{V}|-k$ words where $\mathcal{V}$ represents the whole vocabulary.} which 
excludes a small set of dominating words for the sampling process. Note that we still use the combination of top-k and top-p  sampling to generate easier samples for earlier curriculums. Formally, bottom-k modifies the distribution in Equation \ref{btk} to:
\begin{equation}
\label{rescaled}
P_{\bar{x}_i\in \mathcal{V}\setminus \mathcal{V}^{(k)}}{(\bar{x}_i\mid \bar{x}_1,..., \bar{x}_{i-1}, x)},
\end{equation}
where $\mathcal{V}$ represents the whole vocabulary. Then, at each time step, we sample the next token with the rescaled distribution in Equation \ref{rescaled}. We apply bottom-k for the first $\mathcal{N}$ steps of the generation before fallback to top-k and top-p. Bottom-k tends to generate paraphrases with lower textual similarity. For example, given an input of `I like to remodel homes', existing sampling methods can generate an output `Renovations in property I like to remodel homes'. In contrast, bottom-k sampling generates `Is this what I want to see? Renovating homes are the best choices I have ever had.' where the latter one 
has a higher difficulty measure. Appendix \ref{bottom-kcases} presents an extensive analysis.
\paragraph{Paraphrase Filtering} The inferential properties or mutual implication (MI) has been argued as a form of equivalent meaning \citep{10.2307/4320465, 10.2307/20013377}, i.e., each sentence should entail each other to be `paraphrases'. To support curriculum data augmentation, we exploit mutual implicative paraphrases for grammatical and lexical richness. Algorithm \ref{alg:pcc2} (Lines 5, 6, and 7) depicts the filtering mechanism we propose to generate MI paraphrases. In order to determine the MI relationship between a pair of paraphrase $(x,\bar{x})$, we adopt a pre-trained MI classifier $\mathcal{M}(\cdot,\cdot)$ to calculate a binary indicator $\mathcal{M}(x,\bar{x})$. Here, non-MI paraphrases have a score of $0$ and MI paraphrases have a score of $1$. We also adopt a pre-trained model $\mathcal{G}(\cdot,\cdot)$  to evaluate the textual similarity score of the paraphrases as $\mathcal{G}(x,\bar{x})$. Here, paraphrases with lower similarity scores are treated as grammatically and lexically less similar to the original input sentence. We filter the paraphrase $\bar{x}_i$ based on these two scores:
\begin{equation}
\label{filtering}
\mathcal{M}(x,\bar{x}_i)+(1-\mathcal{M}(x,\bar{x}_i))\mathbbm{1}(\mathcal{G}(x,\bar{x}_i)\geq\beta).
\end{equation}
In the formula above, $\beta$ is a threshold for textual similarity. Here, a paraphrase with a positive mutual implication has a binary output of $1$, i.e., it is preserved regardless of its textual similarity score. A paraphrase with a negative mutual implication  but high textual similarity also has a binary output of $1$, meaning it is preserved as well. In this way, MI paraphrases can be produced. We preserve highly similar paraphrases classified as non-MI, which is a misclassification by the classifier.\footnote{We postulate it as a flaw introduced by the imbalanced training data with a larger portion of paraphrases that tends to be textually unsimilar against the original sentence. We found in our early experiments that removing these easier examples obviously degrades the results for \textsc{Covid-Q} from $51.7$ to $50.0$. Furthermore, ignoring non-MI easy examples prevents PCC from collecting enough augmentation for \textsc{AMZN}.}  All paraphrases that are non-MI with low textual similarity have a binary output of $0$, meaning we discard those paraphrases. After the filtering, a difficulty measure is computed for each paraphrase.\footnote{As in Appendix \ref{impl}, we use an off-the-shelf paraphrase generator and MI classifier in our experiments.}
\paragraph{Difficulty Measure} Recall that for a pair of paraphrase $(x,\bar{x})$, we adopt a pre-trained textual similarity model $\mathcal{G}(\cdot,\cdot)$ to calculate its similarity score as $\mathcal{G}(x,\bar{x})$. BLEURT \citep{sellam-etal-2020-bleurt} score, a BERT-based pre-trained model, is employed as the textual similarity model $\mathcal{G}(\cdot,\cdot)$. Here, paraphrases with lower similarity scores are treated as more difficult instances with higher difficulty measures. For further illustration, we present 6 samples generated from our model in Table \ref{MI_examples} with descending order sorted on the similarity scores.  Here, the similarity scores decently represent the grammatical and lexical difference between the paraphrases candidates, and the mutual implicative paraphrase candidates are grammatically (Sample 2, 3, 4, 5, and 6) and lexically (Sample 1, 2, 3, 4, 5, and 6) rich.
\par
As the distribution of the similarity scores for the paraphrases varies for different inputs, we compute the difficulty measure for a paraphrase $\bar{x}_i$ with its rank in a sorted list of similarity scores, denoted as $\mathrm{sort(\cdot)}$, in descending order among a bag of paraphrase candidates $\mathcal{X}$:
\begin{equation}
\label{diff}
    d_i=  \lceil \mathcal{C}\times \frac{\mathrm{sort}_{\bar{x}_i\in \mathcal{X}}(\mathcal{G}(\bar{x}_i, x))}{|\mathcal{X}|} \rceil,
\end{equation}
where $\mathcal{C}$ represents the total number of curriculum difficulty levels we define, and $|\mathcal{X}|$ represents the total number of paraphrase candidates we have. Here, the paraphrase $\bar{x}_j$ with the highest similarity score, i.e., $\mathcal{G}(x, \bar{x}_j)=\max_{\bar{x}_i\in \mathcal{X}}{(\mathcal{G}(\bar{x}_i, x))}$, has a rank of 1, therefore, $d_j=1$. The paraphrase $\bar{x}_k$ with the lowest similarity score, i.e., $\mathcal{G}(x, \bar{x}_k)=\min_{\bar{x}_i\in \mathcal{X}}{(\mathcal{G}(\bar{x}_i, x))}$, has a rank of $|\mathcal{X}|$, thus $d_k=\mathcal{C}$. Consequently, a larger rank indicates that the paraphrase is more grammatically and lexically different than the original input, and thus belongs to a harder curriculum. We set $d_i=0$ as the easiest difficulty level for the original data.
\subsubsection{Cyclic Curriculum Data Augmentation} \citet{wei-etal-2021-shot} proposed curriculum data augmentation with a gradual course. The training ends after passing the curriculums once. We found that a typical problem called catastrophic forgetting \citep{Kirkpatrick3521} can hamper the performance during such a gradual course, meaning that the model can gradually forget the knowledge learned in an easier course. The augmentation for later curriculums is a subtask of an easier curriculum and can have lexical overlaps. Formally, the input samples $x^{t+1}$ can have overlapping lexical $x^t_i$ which are the same as $x^{t}_j$, where $t$ and $t+1$ represent the curriculum difficulty levels, and $i$ and $j$ represent the word positions in the sentence. Due to catastrophic forgetting, the model can forget what it has learned earlier. Hence, we propose cyclic learning as shown in Algorithm \ref{alg:pcc} to inform the model which skills would be useful later before retrospecting to  easier curriculums with lower difficulties.

\section{Experimental Setup}
In our experiments, we define six curriculums ranging from 0 to 5. 0 represents the original data, and 1 and 5 represent the easiest and the most difficult curriculum respectively.\footnote{We release the code and resource at \url{https://github.com/HongyuanLuke/PCC}.}
\subsection{Few-shot Text Classification Task}
\label{ftc}
For the downstream application task for our experiments, we follow \citet{wei-etal-2021-shot} to conduct the task of few-shot, highly multi-class text classification \citep{10.5555/2627435.2638582,kumar-etal-2019-improving}, which typically has a large number of classes with only a few samples for each of the class. We use triplet loss, a loss computed with three elements, namely, an anchor $a$, a positive sample $p$, and a negative sample $n$. It origins from the vision community \citep{Schroff_2015_CVPR}, which was later applied to language tasks \citep{ein-dor-etal-2018-learning, lauriola2020contextbased}, suitable for the few-shot setting. Precisely, the learning objective is defined as:
\begin{align*}
    \mathcal{L}=\mathrm{\mathcal{D}}(a, p) - \mathcal{D}(a, n) + \gamma,
\end{align*}
where $\mathcal{D}$ represents a distance measure that computes the distance between the input encodings. $\gamma$ represents the margin between the positive and negative samples. We use BERT-based \citep{BERT} pooled sentence encodings as the input into a two-layer triplet network \citep{Schroff_2015_CVPR}.
\par
Three datasets for the text classification task are used in our experiments, namely, \textsc{HUFFPOST} \citep{huff,misra2021sculpting}, \textsc{COVID-Q} \citep{2020arXiv200512522W}, and \textsc{AMZN} \citep{amzn}. For space reasons, we leave their detailed dataset description in Appendix \ref{dstc}.
\subsection{Dialogue Generation Task}
The second downstream task for our experiments is open-domain dialogue generation. We adopt a Seq2Seq neural network \citep{S2S} which receives a text concatenation of prepended knowledge $k$ and dialogue context $c$ and generates the dialogue response $r$ in an autoregressive manner \citep{GPT}. We train our dialogue generator by maximising the following likelihood: \begin{align*}
    P\,(r\mid k, c)=\prod_{t=1}^{T}P\,(r_t\mid r_1,..., r_{t-1}, k, c),
\end{align*}
where $T$ represents the length of the generated dialogue response and $r_t$ represents the word at the position $t$ that has been inferenced. Typical prepended knowledge include personal traits \citep{PERSONACHAT} and movie description \citep{zhou-etal-2018-dataset}. We use DialoGPT \citep{DIALOGPT} for parameter initialization for PCC.
\par
We use \textsc{PersonaChat} (\textsc{ConvAI2}, \citealt{PERSONACHAT}) as the dataset for dialogue generation, which is described in Appendix \ref{dstc2}.
\subsection{Baselines for Text Classification}
\label{baselines1}
We use the following baselines from existing data augmentation methods for text classification.
\paragraph{Triplet Loss} As described in Section \ref{ftc}, an anchor, a positive example and a negative example is selected to construct the loss \citep{Schroff_2015_CVPR}.
\paragraph{Token Substitution} It substitutes words with their WordNet synonyms \citep{zhang2015character,wordnet}.
\paragraph{Pervasive Dropout} It uses dropout on words with probability $p=0.1$ \citep{sennrich-etal-2016-edinburgh}.
\paragraph{SwitchOut} It replaces words with uniformly sampled words \citep{wang-etal-2018-switchout}.
\paragraph{Round-trip Translation} It translates sentences into another language before translating back into the source language \citep{sennrich-etal-2016-improving}.
\paragraph{Hard Negative Mining + EDA} It combines hard negative mining  \citep{Schroff_2015_CVPR} that chooses hard negative samples and EDA \citep{wei-zou-2019-eda} that employs synonym replacement, word-level random insertion, deletion, and swap.
\paragraph{Hard Negative Mining + EDA + Gradual Curriculum} It gradually increases the temperature for EDA augmentation \citep{wei-etal-2021-shot}.
\subsection{Baselines for Dialogue Generation}
\label{baselines2}
We use the following baselines and data augmentation methods for dialogue generation.
\paragraph{TransferTransfo} A Transformer-based model fine-tuned on \textsc{PersonaChat} \citep{2019arXiv190108149W}.
\paragraph{PerCVAE} It uses a memory-augmented architecture with a conditional variational autoencoder to exploit persona information \citep{2019arXiv190512188S}.
\paragraph{DialoGPT} It refers to an autoregressive dialogue generator introduced by \citet{DIALOGPT}.
\paragraph{\textsc{CDA}} It refers to the curriculum data augmentation technique proposed by \citet{wei-etal-2021-shot} using the augmentation of EDA \citep{wei-zou-2019-eda}.
\paragraph{Official \& Flatten} It refers to the paraphrase augmentation technique that is task-specific to the task-oriented dialogue generation \citep{gao-etal-2020-paraphrase}. To adapt it to our task, we use our generated paraphrase via mutual implication, denoted as Flatten, and the official revised \textsc{PersonaChat} paraphrases, denoted as Official.
\paragraph{Round-trip Translation} It translates the input into another language before translating back \citep{sennrich-etal-2016-improving}.
\subsection{Evaluation Metrics}
\label{evalm}
For the text classification task, we follow \citet{wei-etal-2021-shot} to use the top-1 accuracy as the metric.
\par
For the dialogue generation task, we use the word-level F1 score, and we adopt the well-known sequence evaluation metric BLEU \citep{BLEU} where we report BLEU-2, BLEU-3 and BLEU-4. We also adopt another well-known sequence evaluation metric, ROUGE, where we report the F-measures for ROUGE-1, ROUGE-2 and ROUGE-L \citep{rouge}.
\par
To verify our claim that bottom-k sampling generates grammatically and lexically rich paraphrases, we adopt Distinct-$N$ \citep{2015arXiv151003055L, 2018arXiv181105696G} with both $N\in\{1,2,3\}$ and $N\in\{4,5,6\}$ to measure the lexical and grammatical richness respectively using the ratio of distinct $N$-grams against the total number of $N$-grams generated.
\begin{table*}[t!]
\small
\centering
    \setlength\tabcolsep{4pt}
    \setlength\extrarowheight{2pt}
\begin{tabular}{l|ccc|c}
\hline
\noalign{\vskip 1mm}  
\textbf{Model} & \textsc{HuffPost} & \textsc{Covid-Q} &\textsc{AMZN}&  \textbf{Average}\\
\noalign{\vskip 1mm}  
\hline
\hline
\noalign{\vskip 1mm}  
Triplet Loss \citep{Schroff_2015_CVPR}&$20.9 \pm 1.0$& $39.7 \pm 1.0$&$11.6\pm 0.6$ & $24.1$  \\
Triplet Loss + Token Substitution \citep{zhang2015character} &$22.7\pm 1.4$&$43.9\pm 1.3$&$12.8 \pm 0.7$&$26.5$\\
Triplet Loss + Pervasive Dropout \citep{sennrich-etal-2016-edinburgh} & $23.1\pm 1.1$& $43.5\pm 1.8$ & $13.0 \pm 0.6$ & $26.5$\\
Triplet Loss + SwitchOut \citep{wang-etal-2018-switchout} & $22.9 \pm 0.5$ &$41.5\pm 0.6$& $12.7\pm 0.8$ & $25.7$\\
Triplet Loss + Round-trip Translation \citep{sennrich-etal-2016-improving} & $24.2 \pm 0.7$& $42.3 \pm 1.0$ &$13.0 \pm 0.4$& $26.5$\\

Triplet Loss + Hard Negative + EDA \citep{wei-zou-2019-eda} & $22.6 \pm 1.8$ & $48.2 \pm 0.9$ &$13.7\pm 0.9$ & $28.2$\\
\,\,\,\,\,\,\,\,\,\,\,\,\,\,\,\,\,\,\,\,\,\,$\hookrightarrow$ + Gradual Curriculum \citep{wei-etal-2021-shot} & $23.8 \pm 0.9$ & $48.9 \pm 0.9$ & $14.4\pm 1.5$& $29.0$\\
\noalign{\vskip 1mm}  
\hline
\noalign{\vskip 1mm} 
PCC with Cyclic Curr. w/o Bottom-k& $25.2 \pm 1.5$ & $51.4 \pm 0.8$ & $17.4\pm 0.7$& $31.3$\\
PCC with Cyclic Curr. w/ Bottom-k & $\vb*{25.9 \pm 1.7}$ & $\vb*{51.7 \pm 0.6}$ & $\vb*{18.2 \pm 1.0}$& $\vb*{31.9}$\\
\noalign{\vskip 1mm}  
\hline
\end{tabular}
\caption{\label{text_results}
Results in top-1 accuracy for the downstream task of text classification on three datasets. The best results are bolded. We report the results averaged from five random seeds for data selection ranging from 0 to 4, which is the source of the variance here. Our methods report the best performance on all the random data seeds on all the datasets. A combination of top-k and top-p sampling with $k=120$ and $p=0.95$ is used for the penultimate row.}
\end{table*}
\begin{table*}[t!]
\scriptsize
\centering
    \setlength\tabcolsep{1.3pt}
    \setlength\extrarowheight{2pt}
\begin{tabular}{l|c|ccc|ccc}
\hline
\noalign{\vskip 1mm}  
\textbf{Model} & \textbf{F1} & \textbf{BLEU-2} & \textbf{BLEU-3}& \textbf{BLEU-4} & \textbf{ROUGE-1} & \textbf{ROUGE-2} & \textbf{ROUGE-L}\\
\noalign{\vskip 1mm}  
\hline
\hline
\noalign{\vskip 1mm}  
TransferTransfo \citep{2019arXiv190108149W} & $16.61 \pm 0.09$ & $3.16 \pm 0.07$ & $1.04 \pm 0.03$ & $0.43 \pm 0.02$ & $17.69 \pm 0.14$ & $3.96 \pm 0.08$ & $16.34 \pm 0.13$ \\
PerCVAE \citep{2019arXiv190512188S} & $14.33 \pm 0.12$ & $1.23 \pm 0.06$ & $0.20 \pm 0.05$ & $0.04 \pm 0.01$ & $13.25 \pm 0.10$ & $1.62 \pm 0.05$ & $12.02 \pm 0.10$ \\
DialoGPT \citep{DIALOGPT} & $18.58 \pm 0.13$ & $5.25 \pm 0.08$ & $1.89 \pm 0.07$ & $0.66 \pm 0.05$ & $18.42 \pm 0.13$ & $4.62 \pm 0.09$ & $17.23 \pm 0.12$\\
DialoGPT + CDA \citep{wei-zou-2019-eda} & $18.38 \pm 0.10$ & $5.23 \pm 0.10$ & $1.84\pm 0.08$ & 	$0.63\pm 0.02$ & $18.55\pm 0.31$ & $4.63 \pm 0.11$ & $17.40 \pm 0.30$\\
DialoGPT + Flatten \citep{gao-etal-2020-paraphrase} & $18.21\pm 0.21$ & $5.03 \pm 0.18$ & $1.85 \pm 0.11$ & $0.65 \pm 0.04$ & $17.97 \pm 0.34$ & $4.45 \pm 0.16$ & $16.84\pm 0.28$ \\
DialoGPT + Official \citep{gao-etal-2020-paraphrase} & $18.12 \pm 0.11$ & $4.80\pm 0.27$ & $1.78 \pm 0.50$ & $0.59 \pm 0.60$ & $17.88 \pm 0.24$ & $4.38 \pm 0.09$ & $16.84 \pm 0.20$ \\
DialoGPT + RT \citep{sennrich-etal-2016-improving} & $18.26 \pm 0.49$ &$5.10 \pm 0.21$ & $1.80 \pm 0.20$ & $0.62 \pm 0.08$ & $18.32 \pm 0.35$ & $4.47 \pm 0.18$ & $17.16 \pm 0.31$\\
\noalign{\vskip 1mm}  
\hline
\noalign{\vskip 1mm} 
PCC with Cyclic Curr. w/o Bottom-k & $18.76 \pm 0.20$ & $5.38 \pm 0.14$ & $1.99 \pm 0.9$ & $0.71 \pm 0.06$ & $18.81 \pm 0.18$ & $4.75 \pm 0.12$ & $17.53 \pm 0.12$
\\
PCC with Cyclic Curr. w/ Bottom-k & $\vb*{18.80 \pm 0.45}$ & $\vb*{5.59 \pm 0.17}$ &$\vb*{2.07 \pm 0.12}$ &$\vb*{0.76 \pm 0.11}$& $\vb*{19.15 \pm 0.16}$ & $\vb*{4.98 \pm 0.12}$ & $\vb*{17.89 \pm 0.17}$\\
\noalign{\vskip 1mm}  
\hline
\end{tabular}
\caption{\label{dialogue_results}
Results for the downstream task of open-domain dialogue generation on \textsc{PersonaChat}, averaged from three runs. All the metrics attain better quality with higher scores. We denote Round-trip Translation as RT. A combination of top-k and top-p sampling with $k=120$ and $p=0.95$ is used for the penultimate row.}
\end{table*}

\section{Results and Analysis}
\subsection{Few-shot Text Classification Results}
\subsubsection{Main Results}
Table \ref{text_results} presents the results for few-shot text classification. Among the baselines, Triplet Loss + Gradual Curriculum works the best \citep{wei-etal-2021-shot}. PCC improves this baseline significantly. All the models share randomness in data, and our model is the best on all of the random seeds individually. Further, our proposed PCC model surpasses the baselines of Token Substitution, Pervasive Dropout, SwitchOut and Round-trip Translation significantly. Without bottom-k, PCC surpasses all the baselines, and our proposed full model with bottom-k obviously boosts performance. Appendix \ref{analysisDA} additionally presents an analysis of the improvements as a function of the number of data augmentations.
\begin{table}
\scriptsize
\centering
    \setlength\tabcolsep{2pt}
    \setlength\extrarowheight{2pt}

\begin{tabular}{l|ccc}
\hline
\noalign{\vskip 1mm}  
\textbf{Model} & \textsc{HuffPost} & \textsc{Covid-Q} & \textsc{AMZN}\\
\noalign{\vskip 1mm}  
\hline
\hline
\noalign{\vskip 1mm} 
PCC w/o MI filtering & $25.7 \pm 1.4$& $50.2 \pm 1.7$ & $16.7 \pm 1.1$ \\
PCC w/ Pure Sampling & $25.8 \pm 1.0$ & $49.7 \pm 0.9$ & $16.9 \pm 0.8$\\
PCC w/ Inverse Curriculum & $23.0 \pm 1.7$ & $48.5 \pm 1.2$ & $15.0 \pm 0.5$ \\
PCC w/ Random Curriculum& $24.0 \pm 1.7$ & $48.9 \pm 1.5$ & $15.1 \pm 0.8$\\
PCC w/ Gradual Curriculum& $24.7 \pm 1.3$ & $49.6 \pm 1.4$ & $16.5 \pm 0.7$ \\
PCC w/ Inv. Cyc. & $24.9 \pm 1.2$&$50.9 \pm 1.0$ & $16.5\pm 0.8$\\
PCC w/ Cyc. & $25.2\pm 1.5$ & $51.4 \pm 0.8$ & $17.4 \pm 0.7$\\
PCC w/ Inv. Cyc., Bottom-k & $25.3 \pm 1.9$&$51.3 \pm 1.1$ & $17.1 \pm 1.2$\\
PCC w/ Cyc., Bottom-k  & $\vb*{25.9\pm 1.7}$ & $\vb*{51.7 \pm 0.6}$ & $\vb*{18.2 \pm 1.0}$  \\
\noalign{\vskip 1mm}  
\hline
\end{tabular}
\caption{\label{ablation_text}
Ablation results in top-1 accuracy for the downstream task of text classification.
}
\end{table}
\subsubsection{Ablation Study}
Table \ref{ablation_text} presents the results of our ablation study. First, removing the MI paraphrase filtering component described with Equation \ref{filtering} obviously degrades the results. Replacing bottom-k sampling with pure sampling also decreases the results. Furthermore, paraphrasing in a random or an inverse order of decreasing difficulties, i.e., with neither curriculum learning nor cyclic learning, obviously deteriorates the results. \textit{Therefore, our contribution is the discovery of paraphrasing as an effective CDA method rather than using paraphrasing solely as an augmentation technique.} Moreover, using cyclic learning instead of the gradual curriculum improves the results when trained with and without bottom-k sampling. Training the second cycle in an inversed order of decreasing difficulties degrades the results both with and without bottom-k.

\subsection{Analysis on Cyclic Learning}
\label{cyccc}
\begin{figure}[t!]
\begin{center}
\vspace{0mm}
\centerline{
\includegraphics[width=7cm]{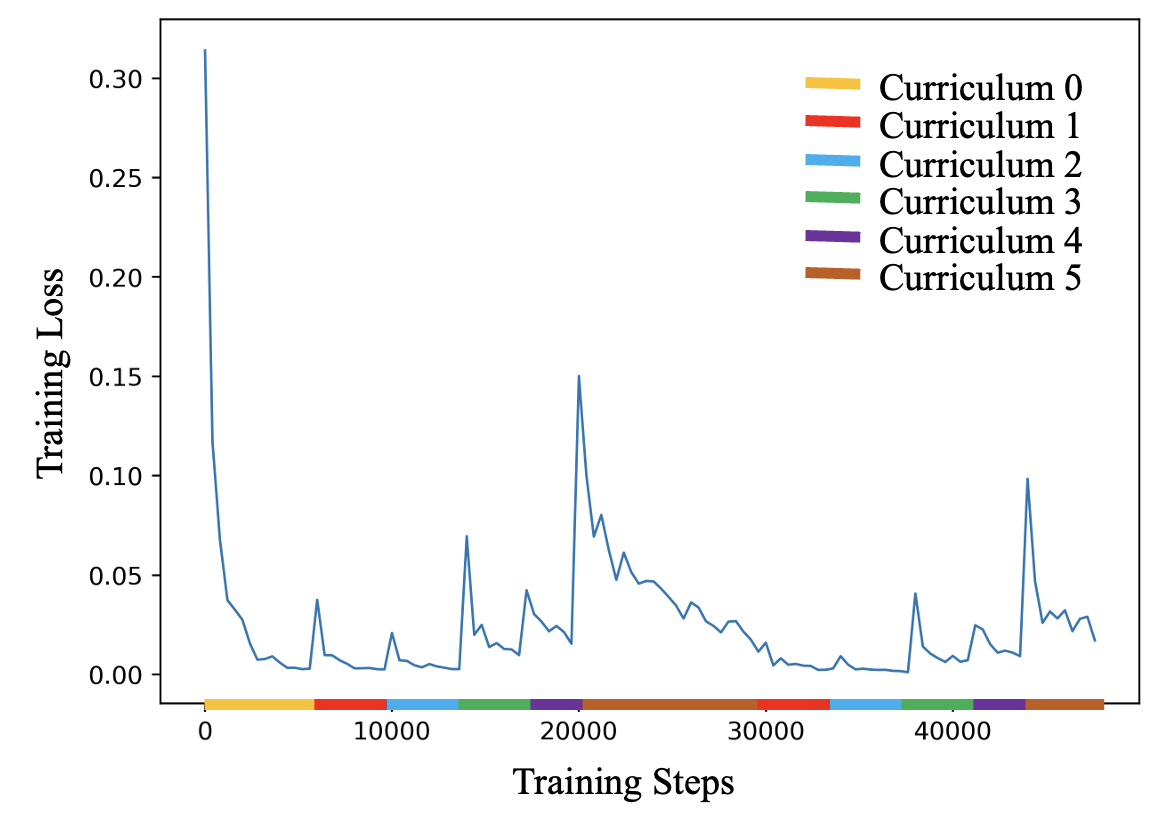}}
\caption{A plot of the training loss for the analysis for cyclic learning. Best viewed in color.}
\label{cyclic_loss}
\end{center}
\vspace{-5mm}
\end{figure}

Figure \ref{cyclic_loss} presents the change of the training loss during the progress of the training on the task of text classification on \textsc{Covid-Q}. We observe that catastrophic forgetting exists as the training loss spikes when re-entering the curriculums. For the second time it enters the most difficult curriculum 5, the loss is also further smoothened compared to the first spike. The spike is also desirable as described in \citet{wei-etal-2021-shot}, indicating that new instances that are harder to learn are presented and can help to escape the local minima. These support the usefulness of our proposed cyclic learning that can smoothen the gradients, mitigate catastrophic forgetting, and improve generalization by entering curriculums multiple times.
\subsection{Dialogue Generation Results}
\label{dkmmmm}
Table \ref{dialogue_results} presents the results for dialogue generation on \textsc{PersonaChat}. First, we present the results for competitive baselines, namely TransferTransfo and PerCVAE. DialoGPT surpasses these two significantly. Using CDA on DialoGPT has deteriorated BLEU scores, which suggests that using CDA causes grammatical influence, possibly due to the random operations that produce undesirable grammatically incorrect augmentation. We also observe a large variance with the official paraphrase provided by \textsc{PersonaChat}, possibly due to the large difference between the manually rephrased sentences. This indicates easier paraphrases seem to be essential for PCC to be effective. Also, the Flatten baseline reported in Table \ref{dialogue_results} approximates a random curriculum, which degrades the results. It leads to a conclusion about the usefulness of the suggested curriculum. Round-trip Translation (RT) seems not effective, which is somehow reasonable as RT was originally designed for machine translation. PCC achieves the best among all the models, suggesting its usefulness for dialogue generation. Appendix \ref{anadia} provides in-depth reasonings on the results. Appendix \ref{human_dialogue_1} presents a human evaluation of the downstream task of dialogue generation. 

\begin{table}[t!]
\scriptsize
\centering
    \setlength\tabcolsep{1pt}
    \setlength\extrarowheight{2pt}
\begin{tabular}{l|cccccc}
\hline
\noalign{\vskip 1mm}  
\textbf{Model} & \textbf{D1} & \textbf{D2} & \textbf{D3} & \textbf{D4} & \textbf{D5} & \textbf{D6}\\
\noalign{\vskip 1mm}  
\hline
\hline
\noalign{\vskip 1mm}  
Pure Sampling & $0.187$ & $0.571$ & $0.788$ & $0.881$ & $0.919$ & $0.932$\\
Top-k\&p ($k$=$120$, $p$=$0.95$)& $0.145$ & $0.481$ & $0.711$ & $0.826$ & $0.877$ & $0.897$\\
Top-k\&p ($k$=$80$, $p$=$0.80$)& $0.125$ & $0.415$ & $0.634$ & $0.762$ & $0.825$ & $0.850$\\
Bot.-k ($k$=2, $\mathcal{N}$=1) & $0.184$ & $0.587$ & $0.824$ & $0.901$ & $0.919$ & $0.925$\\
Bot.-k ($k$=10, $\mathcal{N}$=1) & $0.199$ & $0.630$ & $0.860$ & $0.926$ & $0.940$ & $0.943$\\
Bot.-k ($k$=2, $\mathcal{N}$=5) & $0.223$ & $0.695$ & $0.904$ & $0.945$ & $0.951$ & $0.953$\\
Bot.-k ($k$=5, $\mathcal{N}$=10) & $0.251$ & $0.786$ & $0.950$ & $0.967$ & $0.969$ & $0.970$\\
Bot.-k ($k$=10, $\mathcal{N}$=15) & $\vb*{0.262}$ & $\vb*{0.851}$ & $\vb*{0.971}$ & $\vb*{0.978}$ & $\vb*{0.979}$ & $\vb*{0.979}$\\
\noalign{\vskip 1mm}  
\hline
\end{tabular}
\caption{\label{bottom-kautomatic}
Automatic results for bottom-k sampling on \textsc{PersonaChat}. \textbf{D} represents the Distinct-$N$ scores.
}
\end{table}
\subsection{Analysis on Bottom-k Sampling}
Table \ref{bottom-kautomatic} presents the automatic results for bottom-k sampling on \textsc{PersonaChat}. Here, bottom-k sampling attains the best on Distinct scores with lower grams ($N\in\{1,2,3\}$), indicating its lexical richness. It also attains the best on Distinct scores with higher grams ($N\in\{4,5,6\}$), indicating its grammatical richness. This helps to generate super-hard instances. Note that the setting of bottom-k sampling employed in PCC with $k=2$ and $\mathcal{N}=1$ already gives the best overall diversity against previous sampling methods. Further increasing the value of $k$ and $\mathcal{N}$ leads to higher diversity. 

\subsection{Human Evaluation on Bottom-k Sampling}
\label{humans}
We hired three experienced annotators who have degrees relevant to English Linguistics to conduct an evaluation on bottom-k sampling with \textsc{PersonaChat}. We present a questionnaire composed of 800 questions with 200 randomly sampled training instances with the paraphrases generated with and without bottom-k sampling to the annotators to compare model outputs under A/B testing:
\begin{table}[t!]
    \setlength\tabcolsep{5pt}
    \setlength\extrarowheight{2pt}
\scriptsize
\centering
\begin{tabular}{lcc}

\hline
\noalign{\vskip 1mm}  
\textbf{Criteria} & \textbf{PCC w/o Bottom-k} & \textbf{PCC w/ Bottom-k}\\
\noalign{\vskip 1mm}  
\hline
\hline
\noalign{\vskip 1mm}  
Gramma. Richness & \colorbox{lightgray}{$34$} & \colorbox{cyan}{\textcolor{white}{$\vb*{66}$}}$^{\ddag}$ \\
\noalign{\vskip 1mm} 
Lexical Richness & \colorbox{lightgray}{$33$} & \colorbox{cyan}{\textcolor{white}{$\vb*{67}$}}$^{\ddag}$   \\
\noalign{\vskip 1mm} 
Difficulty & \colorbox{lightgray}{$34$} & \colorbox{cyan}{\textcolor{white}{$\vb*{66}$}}$^{\ddag}$   \\
\noalign{\vskip 1mm} 
Paraphrasing & \colorbox{cyan}{\textcolor{white}{$\vb*{50}$}} & \colorbox{cyan}{\textcolor{white}{$\vb*{50}$}}\textcolor{white}{\dag}   \\
\noalign{\vskip 1mm}  
\hline
\end{tabular}
\caption{\label{humant}
Human evaluation results for bottom-k in winning percentages. $\ddag$ indicates the results as passing a two-tailed binomial significance test with $p < 0.0001$.
}
\end{table}
\begin{itemize}
\setlength\itemsep{0em}
    \item \textbf{(Grammatical Richness)}: \textit{"Which paraphrase do you think is more grammatically different than the original input sentence?"}
    \item \textbf{(Lexical Richness)}: \textit{"Which paraphrase do you think is more lexically different than the original input sentence?"}
    \item \textbf{(Difficulty)}: \textit{"Which paraphrase is more difficult to read and understood?"}
    \item \textbf{(Paraphrasing)}: \textit{"Which one is more like a mutual implicative paraphrase to the input?"}
\end{itemize}
\par
Table \ref{humant} presents the results of our human evaluation. The paraphrases generated by PCC with bottom-k sampling have a significant advantage in lexical and grammatical richness. Such an advantage correlates well with the difficulty of the paraphrases to be understood by human annotators. Furthermore, bottom-k does not hurt the paraphrasing performance compared to the top-k and top-p sampling. The result of human evaluation verifies our claim that bottom-k generates super-hard paraphrases with grammatical and lexical richness. Appendix \ref{bottom-kcases} presents how bottom-k sampling is superior over previous methods in our scenario with case studies about the coping mechanism.
\section{Conclusions}
We propose a novel framework that uses mutual implicative paraphrasing as a curriculum data augmentation technique. Our proposed curriculum-aware paraphrase generation module is composed of three components, a paraphrase candidate generator with a bottom-k sampling strategy for generating superhard paraphrases, a paraphrase filtering mechanism, and a difficulty measure. We propose a bottom-k sampling strategy to effectively generate super-hard instances with grammatical and lexical richness to be used for the later stages in curriculum learning. Moreover, we propose a cyclic learning strategy that mitigates catastrophic forgetting. Experimental results on the task of few-shot text classification as well as dialogue generation support our proposed methodology PCC's usefulness, surpassing several competitive baselines. 
\section*{Limitations}
The proposed PCC cost more computational resources than traditional CDA methods. However, the cost is still affordable. Generating a round-trip augmentation used as one of the baselines costs about 1.5 seconds (1x speed) for \textsc{PersonaChat}. In contrast, generating a single paraphrase costs about 0.40 seconds (3x faster) with PCC on our machine with a single GPU.
\section*{Ethical Statement}
We honour and support the EACL Code of Ethics. The datasets used in this work are well-known and widely used, and the dataset pre-processing does not make use of any external textual resource. In our view, there is no known ethical issue. End-to-end pre-trained dialogue generators are also used, which are subjected to generating offensive context. But the above-mentioned issues are widely known to commonly exist for these models. Any content generated do not reflect the view of the authors.

\bibliography{anthology,custom}

\begin{thebibliography}{57}
\expandafter\ifx\csname natexlab\endcsname\relax\def\natexlab#1{#1}\fi

\bibitem[{Bengio et~al.(2009)Bengio, Louradour, Collobert, and
  Weston}]{BengioLCW09}
Yoshua Bengio, Jérôme Louradour, Ronan Collobert, and Jason Weston. 2009.
\newblock \href {https://doi.org/10.1145/1553374.1553380} {Curriculum
  learning}.
\newblock In \emph{ICML}, pages 41--48.

\bibitem[{Boghossian(1994)}]{10.2307/4320465}
Paul~A. Boghossian. 1994.
\newblock \href {http://www.jstor.org/stable/4320465} {Inferential role
  semantics and the analytic/synthetic distinction}.
\newblock \emph{Philosophical Studies: An International Journal for Philosophy
  in the Analytic Tradition}, 73(2/3):109--122.

\bibitem[{Cai et~al.(2019{\natexlab{a}})Cai, Wang, Bi, Tu, Liu, Lam, and
  Shi}]{cai-etal-2019-skeleton}
Deng Cai, Yan Wang, Wei Bi, Zhaopeng Tu, Xiaojiang Liu, Wai Lam, and Shuming
  Shi. 2019{\natexlab{a}}.
\newblock \href {https://doi.org/10.18653/v1/N19-1124} {Skeleton-to-response:
  Dialogue generation guided by retrieval memory}.
\newblock In \emph{Proceedings of the 2019 Conference of the North {A}merican
  Chapter of the Association for Computational Linguistics: Human Language
  Technologies, Volume 1 (Long and Short Papers)}, pages 1219--1228,
  Minneapolis, Minnesota. Association for Computational Linguistics.

\bibitem[{Cai et~al.(2019{\natexlab{b}})Cai, Wang, Bi, Tu, Liu, and
  Shi}]{cai-etal-2019-retrieval}
Deng Cai, Yan Wang, Wei Bi, Zhaopeng Tu, Xiaojiang Liu, and Shuming Shi.
  2019{\natexlab{b}}.
\newblock \href {https://doi.org/10.18653/v1/D19-1195} {Retrieval-guided
  dialogue response generation via a matching-to-generation framework}.
\newblock In \emph{Proceedings of the 2019 Conference on Empirical Methods in
  Natural Language Processing and the 9th International Joint Conference on
  Natural Language Processing (EMNLP-IJCNLP)}, pages 1866--1875, Hong Kong,
  China. Association for Computational Linguistics.

\bibitem[{Cai et~al.(2020)Cai, Chen, Song, Zhang, Zhao, and
  Yin}]{inproceedings123}
Hengyi Cai, Hongshen Chen, Yonghao Song, Cheng Zhang, Xiaofang Zhao, and Dawei
  Yin. 2020.
\newblock \href {https://doi.org/10.18653/v1/2020.acl-main.564} {Data
  manipulation: Towards effective instance learning for neural dialogue
  generation via learning to augment and reweight}.

\bibitem[{Devlin et~al.(2019)Devlin, Chang, Lee, and Toutanova}]{BERT}
Jacob Devlin, Ming-Wei Chang, Kenton Lee, and Kristina Toutanova. 2019.
\newblock Bert: Pre-training of deep bidirectional transformers for language
  understanding.
\newblock In \emph{NAACL-HLT}.

\bibitem[{Ein~Dor et~al.(2018)Ein~Dor, Mass, Halfon, Venezian, Shnayderman,
  Aharonov, and Slonim}]{ein-dor-etal-2018-learning}
Liat Ein~Dor, Yosi Mass, Alon Halfon, Elad Venezian, Ilya Shnayderman, Ranit
  Aharonov, and Noam Slonim. 2018.
\newblock \href {https://doi.org/10.18653/v1/P18-2009} {Learning thematic
  similarity metric from article sections using triplet networks}.
\newblock In \emph{Proceedings of the 56th Annual Meeting of the Association
  for Computational Linguistics (Volume 2: Short Papers)}, pages 49--54,
  Melbourne, Australia. Association for Computational Linguistics.

\bibitem[{Fan et~al.(2018)Fan, Lewis, and Dauphin}]{topk}
Angela Fan, Mike Lewis, and Yann Dauphin. 2018.
\newblock \href {https://doi.org/10.18653/v1/P18-1082} {Hierarchical neural
  story generation}.
\newblock In \emph{Proceedings of the 56th Annual Meeting of the Association
  for Computational Linguistics (Volume 1: Long Papers)}, pages 889--898,
  Melbourne, Australia. Association for Computational Linguistics.

\bibitem[{Feinerer and Hornik(2020)}]{wordnet}
Ingo Feinerer and Kurt Hornik. 2020.
\newblock \href {https://CRAN.R-project.org/package=wordnet} {\emph{wordnet:
  WordNet Interface}}.
\newblock R package version 0.1-15.

\bibitem[{Ganesh and Corso(2020)}]{Ganesh}
Madan Ganesh and Jason Corso. 2020.
\newblock Rethinking curriculum learning with incremental labels and adaptive
  compensation.

\bibitem[{Gao et~al.(2019)Gao, Bi, Liu, Li, and Shi}]{2018arXiv181105696G}
Jun Gao, Wei Bi, Xiaojiang Liu, Junhui Li, and Shuming Shi. 2019.
\newblock \href {https://doi.org/10.1609/aaai.v33i01.33016383} {Generating
  multiple diverse responses for short-text conversation}.
\newblock \emph{Proceedings of the AAAI Conference on Artificial Intelligence},
  33(01):6383--6390.

\bibitem[{Gao et~al.(2020)Gao, Zhang, Ou, and Yu}]{gao-etal-2020-paraphrase}
Silin Gao, Yichi Zhang, Zhijian Ou, and Zhou Yu. 2020.
\newblock \href {https://doi.org/10.18653/v1/2020.acl-main.60} {Paraphrase
  augmented task-oriented dialog generation}.
\newblock In \emph{Proceedings of the 58th Annual Meeting of the Association
  for Computational Linguistics}, pages 639--649, Online. Association for
  Computational Linguistics.

\bibitem[{Gupta et~al.(2014)Gupta, Bengio, and
  Weston}]{10.5555/2627435.2638582}
Maya~R. Gupta, Samy Bengio, and Jason Weston. 2014.
\newblock Training highly multiclass classifiers.
\newblock \emph{J. Mach. Learn. Res.}, 15(1):1461–1492.

\bibitem[{Holtzman et~al.(2020)Holtzman, Buys, Du, Forbes, and Choi}]{topp}
Ari Holtzman, Jan Buys, Li~Du, Maxwell Forbes, and Yejin Choi. 2020.
\newblock \href {https://openreview.net/forum?id=rygGQyrFvH} {The curious case
  of neural text degeneration}.
\newblock In \emph{International Conference on Learning Representations}.

\bibitem[{Kirkpatrick et~al.(2017)Kirkpatrick, Pascanu, Rabinowitz, Veness,
  Desjardins, Rusu, Milan, Quan, Ramalho, Grabska-Barwinska, Hassabis, Clopath,
  Kumaran, and Hadsell}]{Kirkpatrick3521}
James Kirkpatrick, Razvan Pascanu, Neil Rabinowitz, Joel Veness, Guillaume
  Desjardins, Andrei~A. Rusu, Kieran Milan, John Quan, Tiago Ramalho, Agnieszka
  Grabska-Barwinska, Demis Hassabis, Claudia Clopath, Dharshan Kumaran, and
  Raia Hadsell. 2017.
\newblock \href {https://doi.org/10.1073/pnas.1611835114} {Overcoming
  catastrophic forgetting in neural networks}.
\newblock \emph{Proceedings of the National Academy of Sciences},
  114(13):3521--3526.

\bibitem[{Korbar et~al.(2018)Korbar, Tran, and
  Torresani}]{10.5555/3327757.3327874}
Bruno Korbar, Du~Tran, and Lorenzo Torresani. 2018.
\newblock Cooperative learning of audio and video models from self-supervised
  synchronization.
\newblock In \emph{Proceedings of the 32nd International Conference on Neural
  Information Processing Systems}, NIPS'18, page 7774–7785, Red Hook, NY,
  USA. Curran Associates.

\bibitem[{Kumar et~al.(2019)Kumar, Garg, Mehta, and
  Rasiwasia}]{kumar-etal-2019-improving}
Sawan Kumar, Shweta Garg, Kartik Mehta, and Nikhil Rasiwasia. 2019.
\newblock \href {https://doi.org/10.18653/v1/D19-1604} {Improving answer
  selection and answer triggering using hard negatives}.
\newblock In \emph{Proceedings of the 2019 Conference on Empirical Methods in
  Natural Language Processing and the 9th International Joint Conference on
  Natural Language Processing (EMNLP-IJCNLP)}, pages 5911--5917, Hong Kong,
  China. Association for Computational Linguistics.

\bibitem[{Lauriola and Moschitti(2020)}]{lauriola2020contextbased}
Ivano Lauriola and Alessandro Moschitti. 2020.
\newblock \href {http://arxiv.org/abs/2006.01285} {Context-based transformer
  models for answer sentence selection}.

\bibitem[{Li et~al.(2016)Li, Galley, Brockett, Gao, and
  Dolan}]{2015arXiv151003055L}
Jiwei Li, Michel Galley, Chris Brockett, Jianfeng Gao, and Bill Dolan. 2016.
\newblock \href {https://doi.org/10.18653/v1/N16-1014} {A diversity-promoting
  objective function for neural conversation models}.
\newblock In \emph{Proceedings of the 2016 Conference of the North {A}merican
  Chapter of the Association for Computational Linguistics: Human Language
  Technologies}, pages 110--119, San Diego, California. Association for
  Computational Linguistics.

\bibitem[{{Li} et~al.(2019){Li}, {Weston}, and {Roller}}]{2019arXiv190903087L}
Margaret {Li}, Jason {Weston}, and Stephen {Roller}. 2019.
\newblock \href {http://arxiv.org/abs/1909.03087} {{ACUTE-EVAL: Improved
  Dialogue Evaluation with Optimized Questions and Multi-turn Comparisons}}.
\newblock \emph{CoRR}, abs/1909.03087.

\bibitem[{Lin(2004)}]{rouge}
Chin-Yew Lin. 2004.
\newblock \href {https://www.aclweb.org/anthology/W04-1013} {{ROUGE}: A package
  for automatic evaluation of summaries}.
\newblock In \emph{Text Summarization Branches Out}, pages 74--81, Barcelona,
  Spain. Association for Computational Linguistics.

\bibitem[{Liu et~al.(2018)Liu, He, Liu, and Zhao}]{ijcai2018-587}
Cao Liu, Shizhu He, Kang Liu, and Jun Zhao. 2018.
\newblock \href {https://doi.org/10.24963/ijcai.2018/587} {Curriculum learning
  for natural answer generation}.
\newblock In \emph{Proceedings of the Twenty-Seventh International Joint
  Conference on Artificial Intelligence, {IJCAI-18}}, pages 4223--4229.
  International Joint Conferences on Artificial Intelligence Organization.

\bibitem[{Liu et~al.(2020)Liu, Ren, Tan, Zhang, Qin, Zhao, and
  Liu}]{ijcai2020-534}
Jinglin Liu, Yi~Ren, Xu~Tan, Chen Zhang, Tao Qin, Zhou Zhao, and Tie-Yan Liu.
  2020.
\newblock Task-level curriculum learning for non-autoregressive neural machine
  translation.
\newblock In \emph{Proceedings of the Twenty-Ninth International Joint
  Conference on Artificial Intelligence, {IJCAI-20}}, pages 3861--3867.
  International Joint Conferences on Artificial Intelligence Organization.
\newblock Main track.

\bibitem[{{Miller} et~al.(2017){Miller}, {Feng}, {Fisch}, {Lu}, {Batra},
  {Bordes}, {Parikh}, and {Weston}}]{PARLAI}
Alexander~H. {Miller}, Will {Feng}, Adam {Fisch}, Jiasen {Lu}, Dhruv {Batra},
  Antoine {Bordes}, Devi {Parikh}, and Jason {Weston}. 2017.
\newblock \href {http://arxiv.org/abs/1705.06476} {{ParlAI: A Dialog Research
  Software Platform}}.
\newblock \emph{CoRR}, abs/1705.06476.

\bibitem[{Misra(2018)}]{huff}
Rishabh Misra. 2018.
\newblock \href {https://doi.org/10.13140/RG.2.2.20331.18729} {News category
  dataset}.

\bibitem[{Misra and Grover(2021)}]{misra2021sculpting}
Rishabh Misra and Jigyasa Grover. 2021.
\newblock \emph{Sculpting Data for ML: The first act of Machine Learning}.

\bibitem[{Nighojkar and Licato(2021)}]{nighojkar-licato-2021-improving}
Animesh Nighojkar and John Licato. 2021.
\newblock \href {https://doi.org/10.18653/v1/2021.acl-long.552} {Improving
  paraphrase detection with the adversarial paraphrasing task}.
\newblock In \emph{Proceedings of the 59th Annual Meeting of the Association
  for Computational Linguistics and the 11th International Joint Conference on
  Natural Language Processing (Volume 1: Long Papers)}, pages 7106--7116,
  Online. Association for Computational Linguistics.

\bibitem[{Papineni et~al.(2002)Papineni, Roukos, Ward, and Zhu}]{BLEU}
Kishore Papineni, Salim Roukos, Todd Ward, and Wei-Jing Zhu. 2002.
\newblock \href {https://doi.org/10.3115/1073083.1073135} {{B}leu: a method for
  automatic evaluation of machine translation}.
\newblock In \emph{Proceedings of the 40th Annual Meeting of the Association
  for Computational Linguistics}, pages 311--318, Philadelphia, Pennsylvania,
  USA. Association for Computational Linguistics.

\bibitem[{Peregrin(2006)}]{10.2307/20013377}
Jaroslav Peregrin. 2006.
\newblock \href {http://www.jstor.org/stable/20013377} {Meaning as an
  inferential role}.
\newblock \emph{Erkenntnis (1975-)}, 64(1):1--35.

\bibitem[{Platanios et~al.(2019)Platanios, Stretcu, Neubig, Poczos, and
  Mitchell}]{platanios-etal-2019-competence}
Emmanouil~Antonios Platanios, Otilia Stretcu, Graham Neubig, Barnabas Poczos,
  and Tom Mitchell. 2019.
\newblock \href {https://doi.org/10.18653/v1/N19-1119} {Competence-based
  curriculum learning for neural machine translation}.
\newblock In \emph{Proceedings of the 2019 Conference of the North {A}merican
  Chapter of the Association for Computational Linguistics: Human Language
  Technologies, Volume 1 (Long and Short Papers)}, pages 1162--1172,
  Minneapolis, Minnesota. Association for Computational Linguistics.

\bibitem[{Radford(2018)}]{GPT}
Alec Radford. 2018.
\newblock Improving language understanding by generative pre-training.

\bibitem[{Schroff et~al.(2015)Schroff, Kalenichenko, and
  Philbin}]{Schroff_2015_CVPR}
Florian Schroff, Dmitry Kalenichenko, and James Philbin. 2015.
\newblock Facenet: A unified embedding for face recognition and clustering.
\newblock In \emph{Proceedings of the IEEE Conference on Computer Vision and
  Pattern Recognition (CVPR)}.

\bibitem[{Sellam et~al.(2020)Sellam, Das, and Parikh}]{sellam-etal-2020-bleurt}
Thibault Sellam, Dipanjan Das, and Ankur Parikh. 2020.
\newblock \href {https://doi.org/10.18653/v1/2020.acl-main.704} {{BLEURT}:
  Learning robust metrics for text generation}.
\newblock In \emph{Proceedings of the 58th Annual Meeting of the Association
  for Computational Linguistics}, pages 7881--7892, Online. Association for
  Computational Linguistics.

\bibitem[{Sennrich et~al.(2016{\natexlab{a}})Sennrich, Haddow, and
  Birch}]{sennrich-etal-2016-edinburgh}
Rico Sennrich, Barry Haddow, and Alexandra Birch. 2016{\natexlab{a}}.
\newblock \href {https://doi.org/10.18653/v1/W16-2323} {{E}dinburgh neural
  machine translation systems for {WMT} 16}.
\newblock In \emph{Proceedings of the First Conference on Machine Translation:
  Volume 2, Shared Task Papers}, pages 371--376, Berlin, Germany. Association
  for Computational Linguistics.

\bibitem[{Sennrich et~al.(2016{\natexlab{b}})Sennrich, Haddow, and
  Birch}]{sennrich-etal-2016-improving}
Rico Sennrich, Barry Haddow, and Alexandra Birch. 2016{\natexlab{b}}.
\newblock \href {https://doi.org/10.18653/v1/P16-1009} {Improving neural
  machine translation models with monolingual data}.
\newblock In \emph{Proceedings of the 54th Annual Meeting of the Association
  for Computational Linguistics (Volume 1: Long Papers)}, pages 86--96, Berlin,
  Germany. Association for Computational Linguistics.

\bibitem[{Song et~al.(2019)Song, Zhang, Cui, Wang, and
  Liu}]{2019arXiv190512188S}
Haoyu Song, Wei-Nan Zhang, Yiming Cui, Dong Wang, and Ting Liu. 2019.
\newblock \href {https://doi.org/10.24963/ijcai.2019/721} {Exploiting persona
  information for diverse generation of conversational responses}.
\newblock In \emph{Proceedings of the Twenty-Eighth International Joint
  Conference on Artificial Intelligence, {IJCAI-19}}, pages 5190--5196.
  International Joint Conferences on Artificial Intelligence Organization.

\bibitem[{Su et~al.(2021)Su, Cai, Zhou, Lin, Baker, Cao, Shi, Collier, and
  Wang}]{su-etal-2021-dialogue}
Yixuan Su, Deng Cai, Qingyu Zhou, Zibo Lin, Simon Baker, Yunbo Cao, Shuming
  Shi, Nigel Collier, and Yan Wang. 2021.
\newblock \href {https://doi.org/10.18653/v1/2021.acl-long.137} {Dialogue
  response selection with hierarchical curriculum learning}.
\newblock In \emph{Proceedings of the 59th Annual Meeting of the Association
  for Computational Linguistics and the 11th International Joint Conference on
  Natural Language Processing (Volume 1: Long Papers)}, pages 1740--1751,
  Online. Association for Computational Linguistics.

\bibitem[{Sutskever et~al.(2014)Sutskever, Vinyals, and Le}]{S2S}
Ilya Sutskever, Oriol Vinyals, and Quoc~V. Le. 2014.
\newblock Sequence to sequence learning with neural networks.
\newblock In \emph{Proceedings of the 27th International Conference on Neural
  Information Processing Systems - Volume 2}, NIPS’14, page 3104–3112,
  Cambridge, MA, USA. MIT Press.

\bibitem[{{Vijayakumar} et~al.(2016){Vijayakumar}, {Cogswell}, {Selvaraju},
  {Sun}, {Lee}, {Crandall}, and {Batra}}]{2016arXiv161002424V}
Ashwin~K {Vijayakumar}, Michael {Cogswell}, Ramprasath~R. {Selvaraju}, Qing
  {Sun}, Stefan {Lee}, David {Crandall}, and Dhruv {Batra}. 2016.
\newblock \href {http://arxiv.org/abs/1610.02424} {{Diverse Beam Search:
  Decoding Diverse Solutions from Neural Sequence Models}}.
\newblock \emph{arXiv e-prints}, abs/1610.02424.

\bibitem[{Wang et~al.(2018)Wang, Pham, Dai, and
  Neubig}]{wang-etal-2018-switchout}
Xinyi Wang, Hieu Pham, Zihang Dai, and Graham Neubig. 2018.
\newblock \href {https://doi.org/10.18653/v1/D18-1100} {{S}witch{O}ut: an
  efficient data augmentation algorithm for neural machine translation}.
\newblock In \emph{Proceedings of the 2018 Conference on Empirical Methods in
  Natural Language Processing}, pages 856--861, Brussels, Belgium. Association
  for Computational Linguistics.

\bibitem[{Wei et~al.(2021)Wei, Huang, Vosoughi, Cheng, and
  Xu}]{wei-etal-2021-shot}
Jason Wei, Chengyu Huang, Soroush Vosoughi, Yu~Cheng, and Shiqi Xu. 2021.
\newblock \href {https://doi.org/10.18653/v1/2021.naacl-main.434} {Few-shot
  text classification with triplet networks, data augmentation, and curriculum
  learning}.
\newblock In \emph{Proceedings of the 2021 Conference of the North American
  Chapter of the Association for Computational Linguistics: Human Language
  Technologies}, pages 5493--5500, Online. Association for Computational
  Linguistics.

\bibitem[{Wei and Zou(2019)}]{wei-zou-2019-eda}
Jason Wei and Kai Zou. 2019.
\newblock \href {https://doi.org/10.18653/v1/D19-1670} {{EDA}: Easy data
  augmentation techniques for boosting performance on text classification
  tasks}.
\newblock In \emph{Proceedings of the 2019 Conference on Empirical Methods in
  Natural Language Processing and the 9th International Joint Conference on
  Natural Language Processing (EMNLP-IJCNLP)}, pages 6382--6388, Hong Kong,
  China. Association for Computational Linguistics.

\bibitem[{{Wei} et~al.(2020){Wei}, {Huang}, {Vosoughi}, and
  {Wei}}]{2020arXiv200512522W}
Jerry {Wei}, Chengyu {Huang}, Soroush {Vosoughi}, and Jason {Wei}. 2020.
\newblock \href {http://arxiv.org/abs/2005.12522} {{What Are People Asking
  About COVID-19? A Question Classification Dataset}}.
\newblock \emph{CoRR}, page CoRR:2005.12522.

\bibitem[{Weinshall et~al.(2018)Weinshall, Cohen, and
  Amir}]{pmlr-v80-weinshall18a}
Daphna Weinshall, Gad Cohen, and Dan Amir. 2018.
\newblock \href {https://proceedings.mlr.press/v80/weinshall18a.html}
  {Curriculum learning by transfer learning: Theory and experiments with deep
  networks}.
\newblock In \emph{Proceedings of the 35th International Conference on Machine
  Learning}, volume~80 of \emph{Proceedings of Machine Learning Research},
  pages 5238--5246. PMLR.

\bibitem[{{Wolf} et~al.(2019){Wolf}, {Sanh}, {Chaumond}, and
  {Delangue}}]{2019arXiv190108149W}
Thomas {Wolf}, Victor {Sanh}, Julien {Chaumond}, and Clement {Delangue}. 2019.
\newblock \href {http://arxiv.org/abs/1901.08149} {{TransferTransfo: A Transfer
  Learning Approach for Neural Network Based Conversational Agents}}.
\newblock \emph{CoRR}, abs/1901.08149.

\bibitem[{Wu et~al.(2019)Wu, Lv, Zang, Han, and Hu}]{Wu2019ConditionalBC}
Xing Wu, Shangwen Lv, Liangjun Zang, Jizhong Han, and Songlin Hu. 2019.
\newblock Conditional bert contextual augmentation.
\newblock In \emph{ICCS}.

\bibitem[{Xu et~al.(2020{\natexlab{a}})Xu, Zhang, Mao, Wang, Xie, and
  Zhang}]{xu-etal-2020-curriculum}
Benfeng Xu, Licheng Zhang, Zhendong Mao, Quan Wang, Hongtao Xie, and Yongdong
  Zhang. 2020{\natexlab{a}}.
\newblock \href {https://doi.org/10.18653/v1/2020.acl-main.542} {Curriculum
  learning for natural language understanding}.
\newblock In \emph{Proceedings of the 58th Annual Meeting of the Association
  for Computational Linguistics}, pages 6095--6104, Online. ACL.

\bibitem[{Xu et~al.(2020{\natexlab{b}})Xu, Hu, Jiang, Feng, Wang, Huang, Ju,
  Xiao, and Zhu}]{xu-etal-2020-dynamic}
Chen Xu, Bojie Hu, Yufan Jiang, Kai Feng, Zeyang Wang, Shen Huang, Qi~Ju, Tong
  Xiao, and Jingbo Zhu. 2020{\natexlab{b}}.
\newblock \href {https://doi.org/10.18653/v1/2020.coling-main.352} {Dynamic
  curriculum learning for low-resource neural machine translation}.
\newblock In \emph{Proceedings of the 28th International Conference on
  Computational Linguistics}, pages 3977--3989, Barcelona, Spain (Online).
  International Committee on Computational Linguistics.

\bibitem[{Yang et~al.(2021)Yang, Lam, and
  Li}]{yang-etal-2021-contrastive-representation}
Haoran Yang, Wai Lam, and Piji Li. 2021.
\newblock \href {https://doi.org/10.18653/v1/2021.findings-emnlp.409}
  {Contrastive representation learning for exemplar-guided paraphrase
  generation}.
\newblock In \emph{Findings of the Association for Computational Linguistics:
  EMNLP 2021}, pages 4754--4761, Punta Cana, Dominican Republic. Association
  for Computational Linguistics.

\bibitem[{Yury(2020)}]{amzn}
Kashnitsky Yury. 2020.
\newblock \href {https://doi.org/10.13140/RG.2.2.20331.18729} {Hierarchical
  text classification of amazon product reviews.}

\bibitem[{Zhang et~al.(2020{\natexlab{a}})Zhang, Zheng, Shao, Mao, Xi, and
  Huang}]{zhang-etal-2020-dialogue}
Rongsheng Zhang, Yinhe Zheng, Jianzhi Shao, Xiaoxi Mao, Yadong Xi, and Minlie
  Huang. 2020{\natexlab{a}}.
\newblock \href {https://doi.org/10.18653/v1/2020.emnlp-main.277} {Dialogue
  distillation: Open-domain dialogue augmentation using unpaired data}.
\newblock In \emph{Proceedings of the 2020 Conference on Empirical Methods in
  Natural Language Processing (EMNLP)}, pages 3449--3460, Online. Association
  for Computational Linguistics.

\bibitem[{Zhang et~al.(2018)Zhang, Dinan, Urbanek, Szlam, Kiela, and
  Weston}]{PERSONACHAT}
Saizheng Zhang, Emily Dinan, Jack Urbanek, Arthur Szlam, Douwe Kiela, and Jason
  Weston. 2018.
\newblock \href {https://doi.org/10.18653/v1/P18-1205} {Personalizing dialogue
  agents: {I} have a dog, do you have pets too?}
\newblock In \emph{Proceedings of the 56th Annual Meeting of the Association
  for Computational Linguistics (Volume 1: Long Papers)}, pages 2204--2213,
  Australia. Association for Computational Linguistics.

\bibitem[{Zhang et~al.(2015{\natexlab{a}})Zhang, Zhao, and
  LeCun}]{10.5555/2969239.2969312}
Xiang Zhang, Junbo Zhao, and Yann LeCun. 2015{\natexlab{a}}.
\newblock Character-level convolutional networks for text classification.
\newblock In \emph{Proceedings of the 28th International Conference on Neural
  Information Processing Systems - Volume 1}, NIPS'15, page 649–657,
  Cambridge, MA, USA. MIT Press.

\bibitem[{Zhang et~al.(2015{\natexlab{b}})Zhang, Zhao, and
  LeCun}]{zhang2015character}
Xiang Zhang, Junbo Zhao, and Yann LeCun. 2015{\natexlab{b}}.
\newblock Character-level convolutional networks for text classification.
\newblock In \emph{Advances in neural information processing systems}, pages
  649--657.

\bibitem[{Zhang et~al.(2020{\natexlab{b}})Zhang, Sun, Galley, Chen, Brockett,
  Gao, Gao, Liu, and Dolan}]{DIALOGPT}
Yizhe Zhang, Siqi Sun, Michel Galley, Yen-Chun Chen, Chris Brockett, Xiang Gao,
  Jianfeng Gao, Jingjing Liu, and Bill Dolan. 2020{\natexlab{b}}.
\newblock \href {https://doi.org/10.18653/v1/2020.acl-demos.30} {{DIALOGPT} :
  Large-scale generative pre-training for conversational response generation}.
\newblock In \emph{Proceedings of the 58th Annual Meeting of the Association
  for Computational Linguistics: System Demonstrations}, pages 270--278,
  Online. Association for Computational Linguistics.

\bibitem[{Zhou et~al.(2018)Zhou, Prabhumoye, and
  Black}]{zhou-etal-2018-dataset}
Kangyan Zhou, Shrimai Prabhumoye, and Alan~W Black. 2018.
\newblock \href {https://doi.org/10.18653/v1/D18-1076} {A dataset for document
  grounded conversations}.
\newblock In \emph{Proceedings of the 2018 Conference on Empirical Methods in
  Natural Language Processing}, pages 708--713, Brussels, Belgium. Association
  for Computational Linguistics.

\bibitem[{Zou et~al.(2021)Zou, Liu, Hu, and Zhang}]{2021arXiv210904084Z}
Yicheng Zou, Zhihua Liu, Xingwu Hu, and Qi~Zhang. 2021.
\newblock \href {https://aclanthology.org/2021.emnlp-main.169} {Thinking
  clearly, talking fast: Concept-guided non-autoregressive generation for
  open-domain dialogue systems}.
\newblock In \emph{Proceedings of the 2021 Conference on Empirical Methods in
  Natural Language Processing}, pages 2215--2226, Online and Punta Cana,
  Dominican Republic. Association for Computational Linguistics.

\end{thebibliography}
\bibliographystyle{acl_natbib}
\appendix
\section{Implementation Details}
\label{impl}
For text classification, we use the hyper-parameter settings as in \citet{wei-etal-2021-shot} for the gradual course, and we refer to their paper for the detailed settings. For our cyclic learning, we pass through the curriculums twice. We train the same number of steps for each curriculum as we did in the first pass for our second pass, and the remaining hyper-parameters are kept the same. For Token Substituion, Pervasive Dropout, SwitchOut, and Round-trip Translation, we follow \citet{wei-etal-2021-shot} to use the triplet network as the base model and use a two-stage curriculum for those baselines. Following \citet{wei-etal-2021-shot}, we include 20\% original data whenever augmentation is used.
\par
For dialogue generation, we use \textsc{DialoGPT-small} for parameter initialisation. We use a batch size of 4 and a gradient clip of $0.1$. We use validation patience of $10$ based on the validation loss. We use greedy decoding for all of our experiments. The above settings apply to all our baselines and our proposed model fine-tuned on \textsc{DialoGPT}. We start to apply the augmentation after 130,000 steps for data augmentation methods. We train the first, second, third, fourth, and fifth curriculums with 60,000 steps. For  Official, Flatten, and RT, we perform a two-stage curriculum as described by \citet{wei-etal-2021-shot}. We set $\mathcal{N}$ and $k$ as a small value (typically $\mathcal{N}=1$ and $k=2$) for bottom-k sampling. We perform a cyclic repetition for our proposed method for the same number of steps for each curriculum until early stopped.
\par
During our experiments, we apply data augmentation methods on the entire textual input for text classification, and we apply data augmentation methods on the personas traits for persona-based dialogue generation.
We employ an off-the-shelf pre-trained model for both the paraphrase generator and the MI classifier \citep{nighojkar-licato-2021-improving}.
\par
For all of the datasets, we obtain 20 paraphrases after filtering, and we assign 4 paraphrases \citep{wei-etal-2021-shot} to each of the curriculums we have. We use 2 paraphrases obtained with bottom-k sampling for \textsc{COVID-Q} and we use 4 paraphrases obtained with bottom-k sampling for the remaining datasets.
\par
For our models without bottom-k sampling, we use 20 paraphrases generated with a combination of top-k sampling and top-p sampling with $k=120$ and $p=0.95$ for all of the datasets.
\par
We conduct our experiments for dialogue generation on the \textsc{ParlAI} platform \citep{PARLAI}.

\section{Datasets for Text Classification}
\label{dstc}
\begin{itemize}
    \item The \textsc{HuffPost} dataset is composed of 200k news headlines collected from 2012 to 2018, which is categorized into 41 classes such as politics, entertainment, and travel \citep{huff,misra2021sculpting}. We use all the classes and a 70\%\,/\,30\% train\,/\,test split by class \citep{wei-etal-2021-shot}.
    \item The \textsc{Covid-Q} dataset is composed of 87 classes with several questions per cluster which ask about the same thing \citep{2020arXiv200512522W}. We use the official train\,/\,test split with 3 questions per cluster \citep{wei-etal-2021-shot}. 
    \item The \textsc{AMZN} product review dataset \citep{amzn} categorizes products into given reviews. We consider the use of 318 `level-3' classes with at least 6 samples per product.
\end{itemize}
For the few-shot scenario, we need to set the number of samples in each class, $N_c$, to be used to construct the datasets. We use the setting in \citet{wei-etal-2021-shot} where $N_c=3$ for \textsc{Covid-Q} and $N_c=10$ for \textsc{HuffPost}. We set $N_c=2$ for \textsc{AMZN}.
\begin{table*}[thb!]
\setlength\aboverulesep{0pt}\setlength\belowrulesep{0pt}
\setcellgapes{0pt}\makegapedcells
\scriptsize
\centering
\begin{tabular}{cccccc}
\begin{tabularx}{\textwidth}{X|X|X|X|X|X}
\hline
\noalign{\vskip 1mm} 
\textbf{Sample Number} & $\mathbf{\tau=0.1}$ &$\mathbf{\tau=0.2}$&$\mathbf{\tau=0.3}$&$\mathbf{\tau=0.4}$&$\mathbf{\tau=0.5}$\\
\noalign{\vskip 1mm}  
\hline
\hline
\noalign{\vskip 1mm}    
i) & I equal am glad to help you. & I am glad help to you. & I am gald to happy help you. & To help you. & Glad am to help I you.\\
\noalign{\vskip 1mm}  
\hline \noalign{\vskip 1mm}  
ii) & I am glad you help to. & I am gladiola to help you. & I am glad to assistance you. & Help glad am to i you. & I am gladiolus to helper you.\\
\noalign{\vskip 1mm}  
\hline \noalign{\vskip 1mm}  
iii) & Am glad you. & I am glad help you. & I am glad you help to. & You I gald to help am. & I am glad help you.\\
\noalign{\vskip 1mm}  
\hline
\noalign{\vskip 1mm} 
iv) & I am glad to help you. & I am glad equal to help you. & I am glad to help you. & I am glad to happy happy help you. & I am happy to avail you.\\
\noalign{\vskip 1mm}  
\hline \noalign{\vskip 1mm} 
\end{tabularx}
\end{tabular}
\caption{\label{eda_examples}
Randomly selected cases for an input `I am glad to help you.' using Easy Data Augmentation \citep{wei-zou-2019-eda}. We present recommended temperatures $\tau$ ranging from 0.1 to 0.5, with four samples for each $\tau$.
}
\end{table*}
\section{Dataset for Dialogue Generation}
\label{dstc2}
\textsc{ConvAI2} is an official competition built based on \textsc{PersonaChat} by adding new training examples as well as a hidden test set. For convenience, we denote the former as \textsc{PersonaChat} in the remaining of the paper. Since the test set is not publicly available, we use the official split containing a training\,/\,development split with 8,939\,/\,1,000 multi-turn dialogues conditioned on 1,155\,/\,100 personas respectively. Each persona is composed of about 4 to 5 persona traits.

\begin{table}[t!]
\scriptsize
    \setlength\tabcolsep{1.5pt}
    \setlength\extrarowheight{2pt}
\centering
\begin{tabular}{l|ccccc}
\hline
\noalign{\vskip 1mm}  
\textbf{Model} & $[0.5,$ &  $[0,0.5)$ & $(-0.5,0)$ & $,-0.5]$ & \textbf{Avg.}\\
\noalign{\vskip 1mm}  
\hline
\hline
\noalign{\vskip 1mm}  
Official Paraphrases & $1\%$ & $14\%$ & $33\%$ & $52\%$ & $-0.46$  \\
Round-trip Translation  &  $25\%$ & $52\%$ & $17\%$ & $6\%$ & $0.23$ \\
PCC w/o Bottom-k &  $39\%$ & $11\%$ & $23\%$ & $27\%$ & $0.02$\\
PCC w/ Bottom-k&  $16\%$ & $8\%$ & $18\%$ & $58\%$ & $-0.43$\\
\noalign{\vskip 1mm}  
\hline
\end{tabular}
\caption{\label{ana_dialog}
Analysis on the distribution for the textual similarity score with different augmentation methods.}
\end{table}
\section{Analysis on Dialogue Generation}
\label{anadia}
Table \ref{dialogue_results} reports an ablation when we use our PCC to train the dialogue generator without the use of bottom-k sampling. The results suggest that using bottom-k sampling improves all the metrics, especially the ROUGE scores. Table \ref{ana_dialog} presents the distribution of the textual similarity scores for the paraphrases generated from four methods on \textsc{PersonaChat}. The official paraphrase \citep{PERSONACHAT} largely differs from the original ones, which we postulate as the reason for the large variance observed in Table \ref{dialogue_results}. This also indicates the neccessity of the easier samples for curriculum learning. The Round-trip Translation generates paraphrases that have higher textual similarity with the input sentence. Our method without bottom-k sampling (we use a combination of top-k and top-p sampling with $k=120$ and $p=0.95$ here) generates paraphrases with more evenly distributed scores, with an average of $0.02$. In contrast, bottom-k helps to generate harder samples while still capable of generating more easier samples.
\section{Problematic Cases for EDA}
\label{EDA_problematic}
Table \ref{eda_examples} presents samples from EDA for a sample input `I am glad to help you.' with each of the temperatures $\tau$ ranging from 0.1 to 0.5, which is the recommended setting from \citet{wei-etal-2021-shot}. We categorize EDA's problems as the followings: 
\begin{table*}[thb!]
\tiny
\centering
\begin{tabular}{ccc}
\begin{tabularx}{\textwidth}{X|X|X}
\hline
\noalign{\vskip 1mm} 
\textbf{Original Input Sentence} & \textbf{PCC w/o Bottom-k Sampling }&\textbf{PCC w/ Bottom-k Sampling}\\
\noalign{\vskip 1mm}  
\hline
\hline
\noalign{\vskip 1mm}    
1): i like to shoot a bow. &
When i first started shooting bows, this was the most important method. & Hey, i like to shoot a bow. Just started using a Bow SLR shooter, but a DSLR isn't really necessary.\\
\noalign{\vskip 1mm}  
\hline \noalign{\vskip 1mm}  
2): i have four sisters. & i have four sisters & four sisters, and i want four sisters.\\
\noalign{\vskip 1mm}  
\hline \noalign{\vskip 1mm}  
3): i believe that mermaids are real. & i believe that mermaids are real " @JesseyHawkins & Marxist philosopher,'mermaids are real," property\\
\noalign{\vskip 1mm}  
\hline
\noalign{\vskip 1mm} 
4): i work as a stand up comedian. & jesse t
& trained comedian, I work as a stand up comedian.\\
\noalign{\vskip 1mm}  
\hline
\noalign{\vskip 1mm} 
5): my favorite drink is cuba libre. & My favorite beverage is Cuba libre. & Cuba is my favorite drink and I live in Cuba free.\\
\noalign{\vskip 1mm}  
\hline
\noalign{\vskip 1mm} 
6): i did a few small roles in tv series. & I have done a few small roles in tv series. & over the years i've appeared in a few small roles in television series\\
\noalign{\vskip 1mm}  
\hline
\noalign{\vskip 1mm} 
7): i love bicycling. & bicycle is my friend. i Love Bicycling. & how wonderful \& amp ; inspiring! I love bicycling.\\
\noalign{\vskip 1mm}  
\hline
\noalign{\vskip 1mm} 
8): i own a hearse. & own a hearse. u could do that? & belongs to a hearse. it's not that expensive.\\
\noalign{\vskip 1mm}  
\hline
\noalign{\vskip 1mm}
9): i like to listen to music. & i like to listen to music. How do you make up your mind? & I like to listen to music. by JACK CLINTON\\
\noalign{\vskip 1mm}  
\hline
\noalign{\vskip 1mm}
10): i like to party. & I like to party & touts my ambition and passion for parties " by @MargotHillary by @anadulka @KelisStout\\
\noalign{\vskip 1mm}  
\hline
\noalign{\vskip 1mm}
11): my favorite band is imagine dragons. & my favorite band is imagine dragons. I am just so happy about that. &i love this band it is awesome\\
\noalign{\vskip 1mm}  
\hline
\noalign{\vskip 1mm}

12): i love to sing. & sing, am i love to sing  & artist, i love to sing.\\

\noalign{\vskip 1mm}  
\hline \noalign{\vskip 1mm} 
\end{tabularx}
\end{tabular}
\caption{\label{shi}
Extensive case studies on \textsc{PersonaChat} support our claim that bottom-k sampling generates grammatically and lexically rich paraphrases that are more different than the input sentence.
}
\end{table*}
\begin{itemize}
\setlength\itemsep{0em}
    \item Sample i) with $\tau=0.1$ and sample ii) with $\tau=0.2$ changes the meaning of the input sentence. `equal' is possibly produced by random insertion and `gladiola' is possibly produced by synonym replacement via WordNet \citep{wordnet}.
    \item Most of the samples produced with  $\tau=0.4$ and $\tau=0.5$ breaks the grammar, which can be harmful to generation tasks.
    \item Sample ii) and iv) with $\tau=0.5$ introduces rare words such as `avail' and `gladiolus', which is counterintuitive to see in many tasks.
\end{itemize}
As illustrated in Figure \ref{curriculum_vs}, PCC effectively reduces the above-mentioned issues.
\section{Analysis on Bottom-k Sampling}
\label{bottom-kcases}
\begin{figure}[t!]
\begin{center}
\vspace{0mm}
\centerline{
\includegraphics[width=7cm]{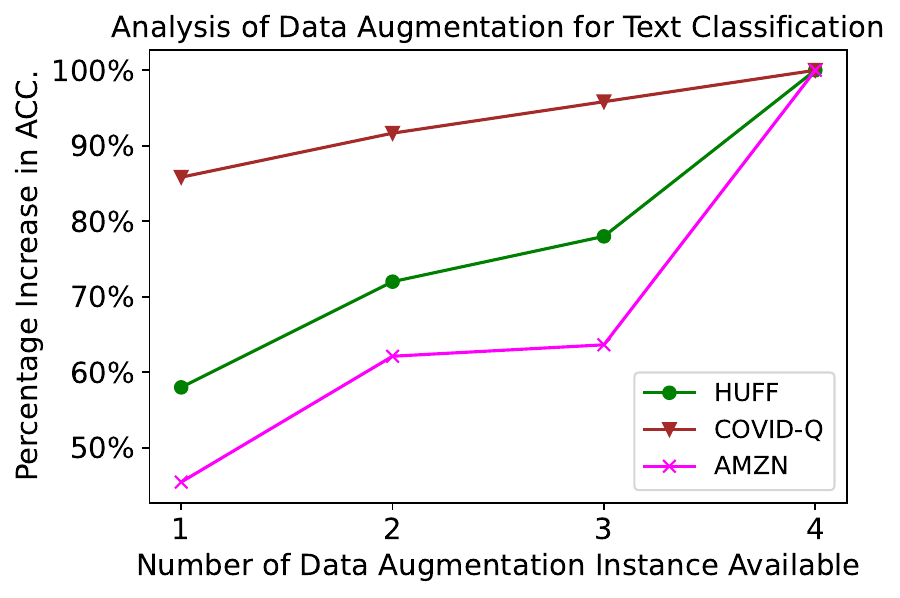}}
\caption{A plot of the percentage performance improvements of the downstream task of text classification against the number of data augmentation instances per curriculum. We use the first row in Table \ref{text_results} as the baseline and the last row in Table \ref{text_results} as the full improvements.}
\label{anada_figure}
\end{center}
\vspace{-5mm}
\end{figure}
Table \ref{shi} presents extensive case studies to support that bottom-k sampling generates grammatically rich and lexically rich paraphrases. PCC without bottom-k tends to exploit a coping mechanism at the beginning of generation (Sample 2, 3, 5, 6, 7, 8, 9, 10, 11, 12). By excluding these dominating words to be copied for generation, bottom-k effectively emphasises the content (Sample 5), improves grammatical richness (Sample 1, 2, 3, 4, 5, 6, 7, 10, 12) and lexical richness (Sample 3, 4, 6, 8, 10, 12), does appropriate synonym replacement (Sample 8, 11) and insertion (Sample 4). Without bottom-k sampling, the input that starts with a first-person pronoun `i' is highly likely to have an output that starts with `i' (Sample 2, 3, 6, 8, 10). This constrains the model from generating grammatically rich paraphrases. In contrast, bottom-k sampling effectively reduces such cases and biases the generation towards a grammatically rich sampling space. Indeed, out of the 6,126 persona traits from \textsc{PersonaChat}, 5,087 of them start with `i'. PCC without bottom-k generates 2,558 paraphrases that start with `i', which avoids generating super-hard instances and hampers the PCC performance.\footnote{Neither a pure top-p sampling with $p=0.95$, a pure top-k sampling with $k=120$, nor a greedy sampling helps, which generates 2,313, 2,381 and 3,302 paraphrases that start with `i' respectively. Compared to these sampling strategies, bottom-k is an effective strategy in preventing copying.} In contrast, bottom-k generates 205 paraphrases that start with `i', indicating its usefulness in improving grammatical richness and generating super-hard instances. Avoiding coping helps to unearth the diverse paraphrases hidden in the tail vocabularies, which we postulate as the reason for the results observed in human evaluation in Section \ref{humans}.
\par

Note that we use bottom-k sampling to effectively prevent coping to generate instances that are textually more different to the input. There is a stream of work that considers improving the diversity \citep{2016arXiv161002424V}. However, these works do not directly consider the similarity between the input paraphrase and the output paraphrase. This is the advantage of bottom-k sampling over this stream of work for our scenario.

\section{Analysis on Data Augmentation}
\label{analysisDA}

Figure \ref{anada_figure} presents the percentage improvements in accuracy as a function of the number of data augmentation instances available for each curriculum. Here, since we have 5 curriculum difficulty levels in our setting, having 3 instances available for each curriculum means that we have 15 data augmentations in total for each original sample. The improvements are positively correlated with the number of available instances. Furthermore, it seems that the improvements of PCC are not saturated yet. This means that a further increase in the number of data augmentations can lead to even higher performance than reported in our paper.
\section{More Human Evaluation}
\label{human_dialogue_1}
\begin{table}[t!]
    \setlength\tabcolsep{15pt}
    \setlength\extrarowheight{2pt}
\scriptsize
\centering
\begin{tabular}{lcc}

\hline
\noalign{\vskip 1mm}  
\textbf{Criteria} & \textbf{w/o PCC} & \textbf{w/ PCC}\\
\noalign{\vskip 1mm}  
\hline
\hline
\noalign{\vskip 1mm}  
Appropriateness & \colorbox{lightgray}{$49$} & \colorbox{cyan}{\textcolor{white}{$\vb*{51}$}}\textcolor{white}{\dag} \\
\noalign{\vskip 1mm} 
Informativeness & \colorbox{lightgray}{$45$} & \colorbox{cyan}{\textcolor{white}{$\vb*{55}$}}$^{\dag}$   \\
\noalign{\vskip 1mm} 
Engagingness & \colorbox{lightgray}{$48$} & \colorbox{cyan}{\textcolor{white}{$\vb*{52}$}}\textcolor{white}{\dag}   \\
\noalign{\vskip 1mm} 
Human-likeness & \colorbox{lightgray}{$49$} & \colorbox{cyan}{\textcolor{white}{$\vb*{51}$}}\textcolor{white}{\dag}   \\
\noalign{\vskip 1mm}  
\hline
\hline
\end{tabular}
\caption{\label{human_dialogue}
Human evaluation results for PCC in winning percentages. $\dag$ indicates the results as passing a two-tailed binomial significance test with $p < 0.05$.
}
\end{table}
\begin{itemize}
\setlength\itemsep{0em}
    \item \textbf{(Appropriateness)}: \textit{"Who is more appropriate given the previous dialogue context?"}
    \item \textbf{(Informativeness)}: \textit{"Who is more diverse instead of null answers such as I do not know?"}
    \item \textbf{(Engagingness)}: \textit{"Who would you prefer to talk with for a long conversation?"}
    \item \textbf{(Human-likeness)}: \textit{"Which speaker do you think sounds more like a real person?"}
\end{itemize}
We follow \citet{2019arXiv190903087L} and \citet{2021arXiv210904084Z} to conduct a human evaluation of dialogue generation from the four aspects described above. We follow the settings used in Section \ref{humans} to invite three experienced annotators to mark 200 instances under A/B settings. The results in Table \ref{human_dialogue} indicate that PCC effectively improves the \textsc{DialoGPT} baseline in all aspects, especially informativeness.

\section{Computing Infrastructure}
\vspace{1mm}
We use an NVIDIA TITAN RTX with 24GB GPU memory for all of the experiments conducted in this paper. 
Training the text classification model consumes about 1 hour. Fine-tuning the dialogue generator consumes about 15 hours. Generating a single paraphrase to be used in PCC as a CDA method costs about 0.40 seconds on our machine.

\end{document}